\documentclass[letterpaper]{article} % DO NOT CHANGE THIS
\usepackage{aaai2026}  % DO NOT CHANGE THIS
\usepackage{times}  % DO NOT CHANGE THIS
\usepackage{helvet}  % DO NOT CHANGE THIS
\usepackage{courier}  % DO NOT CHANGE THIS
\usepackage[hyphens]{url}  % DO NOT CHANGE THIS
\usepackage{graphicx} % DO NOT CHANGE THIS
\usepackage{natbib}  % DO NOT CHANGE THIS AND DO NOT ADD ANY OPTIONS TO IT
\usepackage{caption} % DO NOT CHANGE THIS AND DO NOT ADD ANY OPTIONS TO IT
\usepackage{latexsym}
\usepackage{multirow}
\usepackage{amsmath}
\usepackage{amssymb}
\usepackage{booktabs}
\usepackage{subcaption}
\usepackage{bbding}
\usepackage{colortbl}
\usepackage[T1]{fontenc}
\usepackage{xcolor}         % colors
\usepackage{color}
\usepackage{colortbl,array}
\usepackage{microtype}
\usepackage{url}
\usepackage{booktabs}
\usepackage{multirow}
\usepackage{lineno}
\usepackage{amsmath}
\usepackage{amssymb}   % For \checkmark
\usepackage{pifont}    % For \ding
\usepackage{enumitem}
\usepackage{graphicx} 
\definecolor{darkblue}{rgb}{0, 0, 0.5}

\usepackage{algorithm}
\usepackage{algorithmic}
\definecolor{deepred}{RGB}{180,0,0}

\usepackage{newfloat}
\usepackage{listings}
\DeclareCaptionStyle{ruled}{labelfont=normalfont,labelsep=colon,strut=off} % DO NOT CHANGE THIS
\floatstyle{ruled}
\newfloat{listing}{tb}{lst}{}
\floatname{listing}{Listing}
\title{CO-Bench: Benchmarking Language Model Agents in \\Algorithm Search for Combinatorial Optimization}

\author{
Weiwei Sun\equalcontrib \quad Shengyu Feng\equalcontrib \quad Shanda Li \quad Yiming Yang
}
\affiliations{
    Carnegie Mellon University\\
    \texttt{\{weiweis,shengyuf,shandal,yiming\}@cs.cmu.edu}
}

\usepackage{bibentry}
\begin{document}

\maketitle

\begin{abstract}
Although LLM-based agents have attracted significant attention in domains such as software engineering and machine learning research, their role in advancing combinatorial optimization (CO) remains relatively underexplored. 
This gap underscores the need for a deeper understanding of their potential in tackling structured, constraint-intensive problems---a pursuit currently limited by the absence of comprehensive benchmarks for systematic investigation. 
To address this, we introduce CO-Bench, a benchmark suite featuring 36 real-world CO problems drawn from a broad range of domains and complexity levels. 
CO-Bench includes structured problem formulations and curated data to support rigorous investigation of LLM agents. 
We evaluate multiple agentic frameworks against established human-designed algorithms, revealing the strengths and limitations of existing LLM agents and identifying promising directions for future research. 
CO-Bench is publicly available at \url{https://github.com/sunnweiwei/CO-Bench}.
\end{abstract}

\section{Introduction}

Combinatorial Optimization (CO) is a foundational problem class in computer science and operation research, focused on finding optimal solutions in discrete, structured, and constraint-rich domains. It underpins a wide range of real-world applications, including logistics~\citep{COlogistics}, production planning~\citep{CRAMA1997136}, bioinformatics~\citep{gusfield1997algorithms}, etc. Many CO problems are computationally intractable and classified as NP-hard, making exact solutions impractical at scale. As a result, developing effective algorithms often demands significant domain expertise and manual effort—posing a long-standing challenge in both academic research and industrial applications.

\begin{figure}[t!]
  \centering
  \includegraphics[width=1\linewidth]{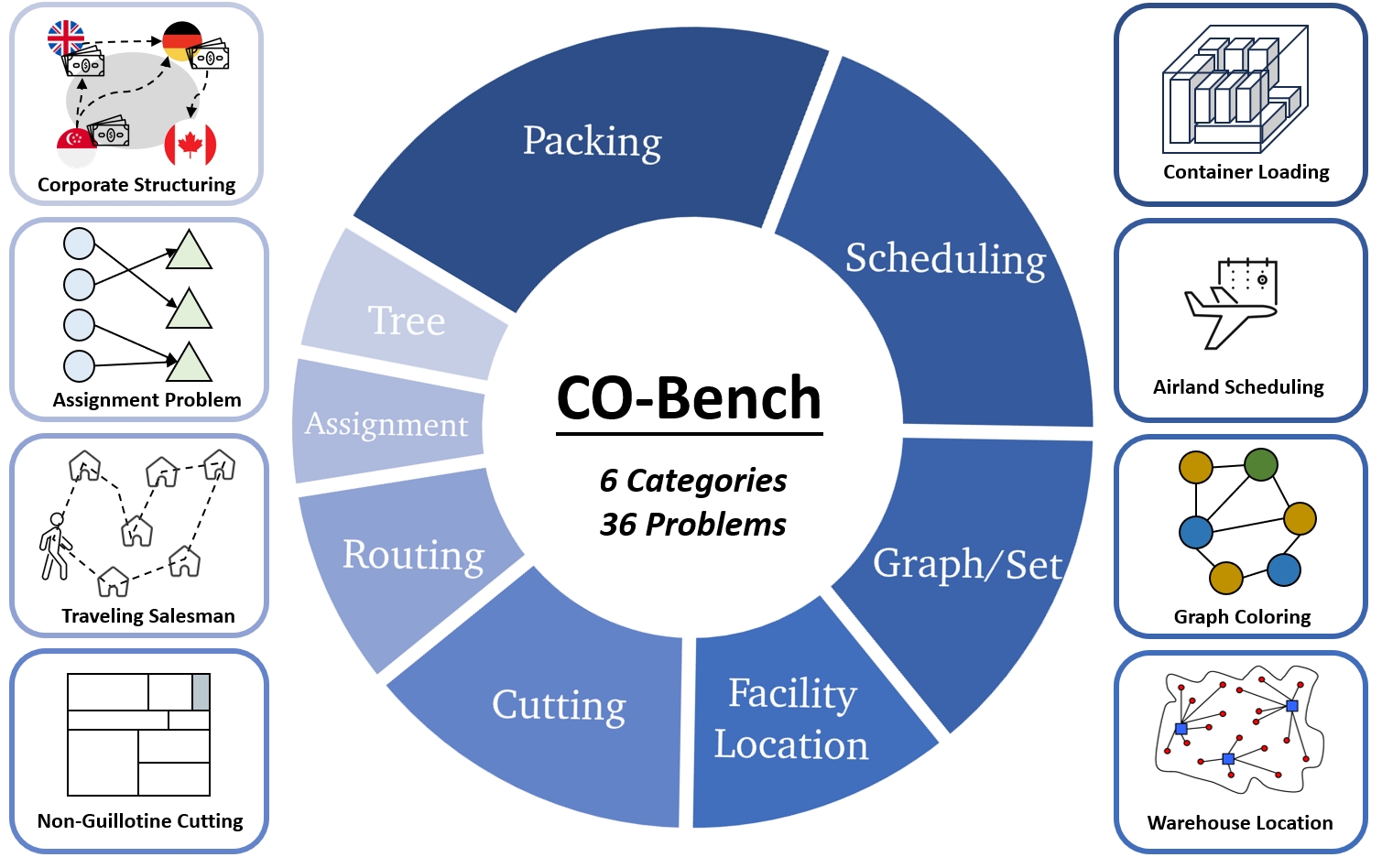} 
  % \vspace{-7mm}
  \caption{Overview of CO-Bench. CO-Bench includes 36 problems from 8 categories, and aims to evaluate LLM agents’ ability to develop effective and efficient algorithms for solving real-world combinatorial optimization problems.}
  \label{fig:example}
\end{figure}

Recent advances in Large Language Models (LLMs)~\citep{openai2024openaio1card, deepseekai2025deepseekr1incentivizingreasoningcapability} have positioned LLM-based agents as increasingly promising tools for a variety of prediction and decision-making tasks~\citep{Jimenez2023SWEbenchCL,Chan2024MLEbenchEM,Gottweis2025TowardsAA}. 
In particular, there is growing interest in applying LLMs to CO problems. 
Initial investigations have largely focused on solution correctness, evaluated on small-scale test instances~\citep{Ramamonjison2023NL4OptCF,YANG2024OptiBenchMR,Xiao2024ChainofExpertsWL}, and are often geared towards solving problems posed by general users.
More recent works have begun to explore autonomous LLM agents capable of conducting research and designing more efficient algorithms for complex scientific and industrial challenges. 
For example, FunSearch~\citep{RomeraParedes2023MathematicalDF} combines LLM prompting with evolutionary search to discover heuristics that outperform human-designed counterparts in the Cap Set and Bin Packing problems. Subsequent methods~\citep{Liu2024EoH,ye2024reevo,Novikov2025AlphaEvolveAC} further improve computational efficiency and broaden applicability to domains such as routing and scheduling.

Despite these advancements, most existing efforts focus on narrow components (e.g., priority functions) within established algorithms, across a limited set of tasks (typically 4-7 problems), and often rely on heavily handcrafted, problem-specific prompts and templates~\citep{RomeraParedes2023MathematicalDF,ye2024reevo}. Furthermore, there remains a lack of systematic evaluation of how these agents perform across a broader and more diverse collection of real-world CO problems.

To address this gap, we introduce \textbf{CO-Bench}, a comprehensive benchmark designed to evaluate LLM agents in the context of efficient CO algorithm development. 
CO-Bench comprises real-world CO problems spanning a wide range of domains and complexities. 
Figure \ref{fig:example} illustrates the problem categories and examples, while Table \ref{table:data} compares CO-Bench with existing CO benchmarks. 
Compared to prior benchmarks, CO-Bench offers broader problem coverage, and emphasizes end-to-end evaluation of LLM-based research agents, focusing on their ability to design efficient, potentially novel algorithms from \textit{abstract problem descriptions}. 
This design enables reproducible and scalable evaluation of agent performance, including comparisons with human-designed classical CO solver under equivalent time constraints. 
By doing so, CO-Bench introduces new challenges for LLM agent development, such as the discovery of algorithms that extend beyond current human knowledge of CO.

Using CO-Bench, we benchmark 15 LLMs and 9 agentic frameworks, comparing their performances against both human-designed classical algorithms and the best-known solutions reported in the literature. 
Our results show that reasoning models (e.g., o3-mini and Claude-3.7-sonnet) consistently outperform standard no-reasoning LLMs. 
When integrated into agentic frameworks like FunSearch, LLMs further improve through trial-and-error exploration. Notably, on 25 problems, LLM-generated algorithms outperformed classical solvers, and on 3 problems, they surpassed the best-known solutions. 
However, our analysis also reveals current limitations, such as limited algorithmic novelty and insufficient handling of feasibility constraints. These findings highlight both the promise and challenges of LLM-driven research in CO and suggest key directions for advancing autonomous algorithm design.

In summary, this paper makes the following contributions: 
\begin{itemize}
    \item[(i)] We introduce CO-Bench, the first comprehensive benchmark to evaluate the capability of LLMs to develop algorithms for diverse and challenging real-world CO problems
    \item[(ii)] We benchmark 15 LLMs and 9 agentic frameworks, analyzing their performance relative to expert-designed pipelines. Our results highlight the strengths of agent-generated algorithms, while also revealing limitations in planning, feasibility checking, and the generation of efficient solution.
\end{itemize}

\section{Preliminary}
\subsection{Combinatorial Optimization}
For each CO problem $c$ (for example, Traveling salesman problem), we follow \citet{christos1982co} to formulate it as a constrained optimization problem in the discrete space. Consider an instance $p$, the optimization problem could be expressed as
\begin{equation}
    \min_{x\in S_{c}(p)} f_c(x; p) + g_c(x; p),
\end{equation}
where $S_{c}(p)$ represents the solution space, e.g., $\mathbf{Z}^m\times \mathbb{R}^{n}$ for $d$ discrete variables and $n$ continuous variables, $f_c(x; p)$ corresponds to the objective function, and $g_c(x; p)$ stands for the constraint violation, which is $0$ for feasible solutions and $+\infty$ otherwise. To avoid the clutter, we simply denote $h_c(x; p) = f_c(x; p) + g_c(x; p)$ in the following text and omit $c$ if the context is clear.

Given an algorithm set $\mathcal{A}$ and a problem instance distribution $D$, the algorithm search problem is  defined as 
\begin{equation}
    \min_{A\in\mathcal{A}} \mathbb{E}_{p\sim D, x\sim A(p)}[h(x; p)].
\end{equation}
In contrast to previous neural CO solvers \citep{bengio2020machine} that directly parameterize $A$ with a neural network, we focus on symbolic searching space where $\mathcal{A}$ consists of all algorithms that could be represented by a Python Program, with a maximum number of $d$ tokens, where $d$ is typically decided by the output length limit of an LLM. Considering the popularity of randomized algorithms \citep{Rajeev2013RA} for CO, we treat the output of an algorithm $A(p)$ as a distribution of solutions, while deterministic algorithms would correspond to the point distributions.

The main endeavor of this work is focused on the shaping of the algorithm set $\mathcal{A}$, the curation of the data distribution $D$ and the definition of $h$ on our collected CO problems.

\begin{table}[t!]
\centering\setlength{\tabcolsep}{2pt}
% \begin{tabular}{l cccccc}
% \toprule
% \textbf{Dataset} & \textbf{CO-Bench} & \textbf{NPHardEval} & \textbf{NL4OPT}  & \textbf{OptiBench} &\textbf{ComplexOR} &  \textbf{ReEvo}  \\
% \midrule

% Algorithm Development & \ding{51} & \ding{55} & \ding{55} & \ding{55} & \ding{55} & \ding{51}\\

% \# of problem types & \textbf{36} & 9 & 5 & 4 & 20$^{\star}$ & 7\\
% \# of test-set instances %test cases 
%     & \textbf{6,482} & 900 & 289 & 605 & 100 & 597 \\
% \# of variables (max) & \textbf{11,000} & 24 & 3 & 18 & 9$^{\star}$ & 1,000 \\

% % Avg \# of test cases per problem & 180 \\
% \bottomrule
% \end{tabular}
\begin{tabular}{l cccc}
\toprule
\multirow{2}{*}{\textbf{Dataset}} &
\multirow{2}{*}{\parbox{1.6cm}{\centering\textbf{Algorithm Dev}}} &
\multirow{2}{*}{\parbox{1.4cm}{\centering\textbf{Problem Num}}} &
\multirow{2}{*}{\parbox{1.4cm}{\centering\textbf{Instance Num}}} &
\multirow{2}{*}{\parbox{1.5cm}{\centering\textbf{Largest Variables}}} \\
&\\
% \textbf{Dataset} & \textbf{Algo Dev?} & & \multirow{2}{*}{\parbox{1.7cm}{\centering\textbf{Valid Solution}}}  & \textbf{Instances Num} & \textbf{Max Var} \\
\midrule
NPHardEval        & \ding{55} & 9           & 900            & 24              \\
NL4OPT            & \ding{55} & 5           & 289            & 3               \\
OptiBench         & \ding{55} & 4           & 605            & 18              \\
ComplexOR         & \ding{55} & 20 & 100            & 9     \\
ReEvo             & \ding{51} & 7           & 597            & 1,000           \\
\midrule
CO\text{-}Bench   & \ding{51} & \textbf{36} & \textbf{6,482} & \textbf{11,000} \\
\bottomrule
\end{tabular}
\caption{Data statistics for CO-Bench and related CO benchmarks, including the indicator for  algorithm development support, the number of problem types, the number of test-set problem instances, and the largest number of test-set variables (e.g., the number of nodes in the largest graph).
}\label{table:data}
\end{table}

\subsection{LLM Agents}
Given a CO problem $c$, a candidate algorithm could be generated by an LLM as
\begin{equation}
A \sim M(\text{textify}(c); \theta),
\end{equation}
where $M$ denotes an LLM with parameters $\theta$. However, one-time generation usually leads to inexecutable code or suboptimal algorithms  \citep{Madaan2023SelfRefineIR}, and \textit{agentic frameworks} address this by enabling iterative refinement through structured interactions with external tools (e.g., a coding environment). Formally, an agent performs reasoning-action iterations~\citep{Yao2022ReActSR}:
\begin{align}
r_{t+1} & \sim M(\text{textify}_r(c, A_t, H_t); \theta),\\
a_{t+1} & \sim M(\text{textify}_a(r_{t+1}, H_t); \theta),
\end{align}
where $r_t$ is the reasoning step, $a_t$ is the action step (e.g., executing code, evaluating results), and $H_t = {(r_i, a_i, \text{result}(a_i))}_{i=1}^{t-1}$ maintains the interaction history.
Thus, an \textit{LLM agent} is formally defined as an LLM $M(\cdot; \theta)$ guided by a structured workflow specifying iterative external interactions to enhance its outputs.

% Large language models can search over space $\mathcal{A}$ 

% Language models are able to generate an algorithm $A$  

% Existing CO solver rely on human / training data / human discovered heuristic, 

% LLM agent, code generation, pipeline, 

\section{CO-Bench}
We introduce CO-Bench, a comprehensive benchmark designed to evaluate the algorithm development ability of LLM agents on combinatorial optimization (CO) problems. The benchmark consists of 36 problems mainly sourced from OR-Library~\citep{Beasley1990ORLibraryDT}, an established archive containing datasets accumulated by researchers across over 30 years of operations research. These problems span a wide range of realistic CO challenges in  academia and industrial applications.

\subsection{Data Curation}

\paragraph{Problem Selection}  We first perform rigorous filtering and cleaning, and select 36 CO problems that cover diverse domains and complexities, including:
\begin{itemize}[ leftmargin=2em]\small
    \item \textit{Packing problems:} Bin packing~\citep{Falkenauer1996AHG}, Multi-Demand Multidimensional Knapsack problem~\citep{Cappanera2001ALH}, Multidimensional knapsack problem~\citep{Petersen1967ComputationalEW}, Container loading~\citep{Bischoff1995IssuesIT,Ivancic1988AnIP}, Container loading with weight restrictions~\citep{Ratcliff1998AllowingFW,Bischoff2006ThreedimensionalPO}, Packing unequal circles~\citep{Lpez2016AFS}, Packing unequal rectangles and squares number / area~\citep{Lpez2018PackingUR}.
     \item \textit{Cutting problems:} Assortment problem~\citep{Beasley1985AnAF}, Constrained / unconstrained guillotine cutting~\citep{Christofides1977AnAF, Beasley1985AlgorithmsFU}, Constrained non-guillotine cutting~\citep{Beasley1985AnET,Beasley2004APH}.
     \item  \textit{Facility location problems:} Capacitated / Uncapacitated warehouse location~\citep{Beasley1988AnAF,Beasley1993LagrangeanHF}, Capacitated / Uncapacitated p-median problem ~\citep{Beasley1985ANO,Osman1994CapacitatedCP}. 
    \item \textit{Scheduling problems:} Aircraft landing~\citep{Beasley2000SchedulingAL,Beasley2004DisplacementPA}, Crew scheduling~\citep{Beasley1996ATS}, Common due date scheduling~\citep{Biskup2001BenchmarksFS}, Flow shop scheduling~\citep{taillard1993benchmarks}, Hybrid Reentrant Shop Scheduling~\citep{Chakhlevitch2009SchedulingRJ}, Job shop scheduling~\citep{taillard1993benchmarks}, Open shop scheduling~\citep{taillard1993benchmarks}.
    \item \textit{Routing problems:} Traveling salesman problem~\citep{LAPORTE1992231}, Period vehicle routing problem ~\citep{Christofides1984ThePR}, Resource constrained shortest path~\citep{Beasley1989AnAF}.
    \item \textit{Assignment problems:} Constrained / unconstrained assignment~\citep{Osman1995HeuristicsFT,Beasley01021990}.
    % Unconstrained assignment~\citep{Beasley01021990} 
    \item \textit{Tree problems:} Euclidean Steiner~\citep{Beasley1992AHF}, Corporate structuring~\citep{Anken2012CorporateSO}
    \item \textit{Graph and set problems:}  Maximal Independent Set~\citep{Erdos1984OnTE},  Graph colouring~\citep{Fleurent1996GeneticAH}, Equitable partitioning~\citep{Mingers1995CreatingSG}, Set partitioning~\citep{Chu1998ConstraintHI},  Set covering~\citep{Beasley1992EnhancingAA}.
\end{itemize}

\paragraph{Data Annotation}
For each problem, we manually annotate the following components:
(1) \textit{Problem description}: a formal definition of the optimization problem in natural language, accompanied by a clearly specified \texttt{solve} function as  the starter code;
(2) \textit{Data loading function}: a \texttt{load\_data} function to load and preprocess raw data from the test files;
(3) \textit{Evaluation function}: an \texttt{eval\_func} function that rigorously and robustly evaluates the quality of a solution.
Additionally, each problem comprises a \textit{development} set and a \textit{test} set, each containing several problem instances.

\begin{figure*}
    \centering
    \includegraphics[width=0.9\linewidth]{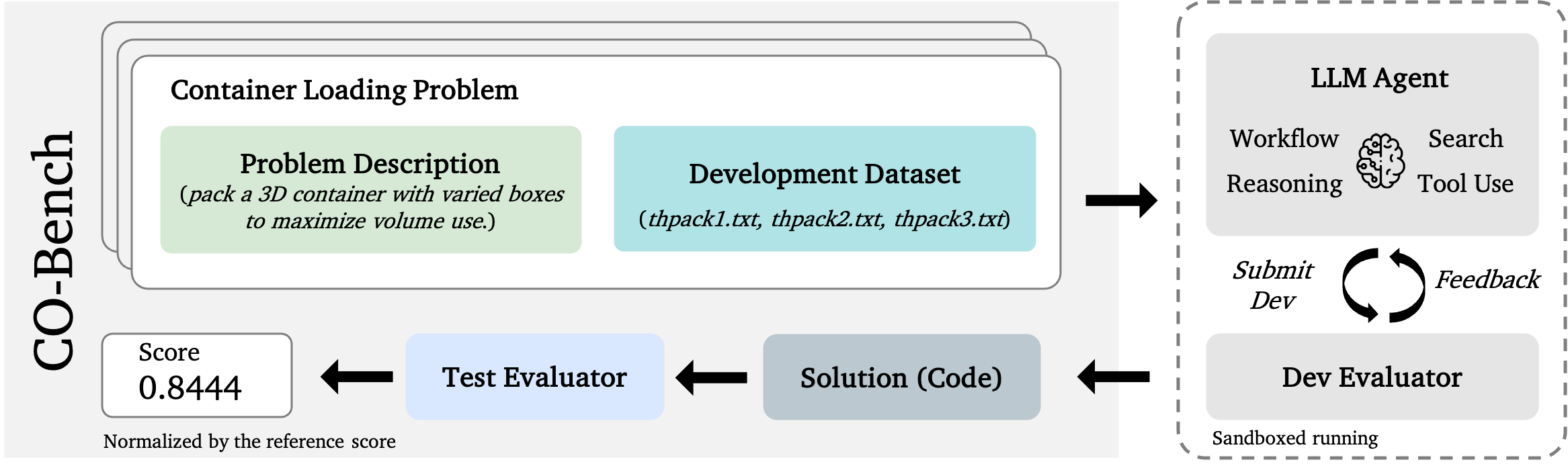}
    \caption{CO-Bench is an evaluation environment for AI agents. Each problem has an associated description and a development dataset. % Agent final solution is code
    Following the setup in \citet{Chan2024MLEbenchEM}, the agent-generated code implements an algorithm design, which is further graded and compared against the best-known solution and human expert solution.}
    \label{fig:evaluation}
\end{figure*}

\paragraph{Evaluation Framework}
We develop a rigorous and efficient evaluation framework to assess the performance of LLM agents in simulated, time-constrained competition scenarios~\citep{Chan2024MLEbenchEM}. Specifically, LLM agents operate within a sandbox environment with access to a Linux machine. For each problem, agents are provided with a problem description, development datasets, and an API endpoint for submitting their solutions (i.e. codebases) to receive evaluation feedback.
An independent evaluation system, which is protected by built-in safeguards, scores the submitted solutions on the development set in parallel.
After a limited number of research steps, the agent submits its  final solution for evaluation on the test set.
During the agent development process, both \texttt{eval\_func} and test data are invisible.
Figure \ref{fig:evaluation} shows the evaluation pipeline in CO-Bench.

\paragraph{Designing Classical Solver Baselines}\label{sec:human}
To investigate how existing LLM agents perform compared to classical solvers, we establish a \textit{classical solver} baseline.
Specifically, the authors of this paper—who have extensive experience in related areas and are familiar with the problems in CO-Bench—spent approximately 30 minutes per problem testing and selecting the most effective classical solvers (e.g., LKH for TSP, CPLEX for scheduling, Gurobi for MIS) and tuning their hyperparameters on the development set.
This process ensures that the classical solver baseline is well-tuned and competitive for each problem in CO-Bench.

\subsection{Evaluation Metrics}
% For comparing performance of different systems, we designed the following metrics.

\paragraph{Avg Score}
The main evaluation metric is similar to the \textit{Primal Gap}~\citep{Berthold2006PrimalHF}, defined as the normalized score of the primal bound \( h(x; p) \) against a pre-computed optimal (or best-known) objective value \( h^*_p \):
\begin{equation}
    s(x,p) = \frac{\min\{|h(x,p)|, |h_p^*|\}}{\max\{|h(x,p)|, |h_p^*|\}},
\end{equation}
A higher value indicates better performance and a score of \(1\) signifies the performance identical to the optimal or best-known solution.
Program errors or infeasible solutions lead to a score of \(0.0\).
The score of a solver on a given problem is computed by averaging its scores across all test instances. The overall benchmark score is then obtained by averaging these problem-level scores across all 36 problems.

\paragraph{Valid Solution}
We compute the percentage of problems for which the generated code is correct on all test instances. Any raised error--such as constraint violation or timeout--is treated as an invalid signal. If any test instance for a given problem results in an invalid signal, the entire solution for that problem is considered invalid, even if it produces valid results on other test instances.

\paragraph{Above Classical}
Given the performance of classical solver, we calculate the portion of problems where the model outperforms the classical solver baseline.

\paragraph{Survival Rate}
The survival rate measures that, for each problem, the percentage of test instances where the model's solution is above 99\% of the reference score (reported optimal or best-known solution from literature). This serve as a challenge metric as the model can only get credit when it is very close or better than previous-best algorithm.

\section{Experimental Setup}

\subsection{Benchmarked Methods}
On CO-Bench, we evaluate various LLMs combined with different agentic frameworks, and compare them with existing human-designed CO solvers.

\paragraph{LLMs}
We conduct experiments on 5 open-source models and 10 proprietary models. These include instruction-tuned models such as \textit{Llama-3.3-70B-Instruct}~\citep{Dubey2024TheL3}, \textit{Qwen-2.5-Code-32B-Instruct}~\citep{Hui2024Qwen25CoderTR}, \textit{DeepSeek-V3}~\citep{DeepSeekAI2024DeepSeekV3TR}, and \textit{GPT-4o}~\citep{Hurst2024GPT4oSC}, as well as frontier reasoning models, including \textit{o3-mini}~\citep{ZhangOpenAIOS}, \textit{Claude-3.7-Sonnet-Thinking}~\citep{claude_sonnet}, \textit{DeepSeek-R1}~\citep{DeepSeekAI2025DeepSeekR1IR}, \textit{Grok-3-Thinking}~\citep{grok3}, \textit{QwQ-32B}~\citep{qwq_32b}, and \textit{Gemini 2.5 Pro}~\citep{gemini_flash_thinking}.

\paragraph{Agentic frameworks}
For the aforementioned LLMs, we apply various agentic frameworks to evaluate their performance across different strategies. These range from simple approaches, such as direct generation, to more sophisticated frameworks that augment LLM with additional tools, workflows, and test-time compute:
\begin{itemize}
    \item \textit{Direct Answer:} The simplest approach, where the LLM directly generates a solution to the combinatorial optimization problem without further refinement.
    \item \textit{BestOfN Sampling~\citep{Chen2021EvaluatingLL}:} Generate $N$ candidate solutions, evaluate each on a development set, and select the solution with the best performance.
    \item \textit{Chain of Experts~\citep{Xiao2024ChainofExpertsWL}:} A multi-agent prompting framework where agents of different roles cooperate to debug and deliver one solution.
    \item \textit{Greedy Refinement~\citep{Shinn2023ReflexionLA,Madaan2023SelfRefineIR}:} Iteratively prompt the LLM to refine the current best solution based on the evaluation results of the development set, repeating this refinement process for $N$ steps.
    \item \textit{FunSearch~\citep{RomeraParedes2023MathematicalDF}:} Prompt the LLM to either draft a new solution or refine an existing one, followed by employing an evolutionary algorithm to iteratively select and improve candidate solutions.
    \item \textit{EoH~\citep{Liu2024EoH}:} 
    Evolve both thoughts and codes in an evolutionary search framework for generating high-performance heuristics. 
    \item \textit{AIDE~\citep{Jiang2025AIDEAE}:} A representative method for machine learning engineering tasks, which stores existing solutions in a tree structure and selectively prompts the LLM to draft new solutions, debug or improve previously stored solutions.
    \item \textit{ReEvo~\citep{ye2024reevo}:} A recent evolutionary algorithm that incorporates short-term and long-term reflection modules, as well as a multi-agentic framework.
    \item \textit{MSTC-AHD~\citep{Zheng2025MonteCT}:} A Monte Carlo Tree Search (MCTS)-based agentic pipeline that organizes all LLM-generated heuristics in a tree structure and uses the MCTS algorithm with progressive widening technique to guide the evolution of heuristics.
    % \item \textit{ReAct:}
    
\end{itemize}

% \begin{figure*}[!ht]
%     \centering
% \includegraphics[width=\linewidth]{figs/all-score.png}
%     \caption{\textbf{Avg Score} of LLMs and Agents on CO-Bench.}
%     \label{fig:benchmark}
% \end{figure*}
% \input{table/main}

\begin{figure*}[!ht]
    \centering
\includegraphics[width=\linewidth]{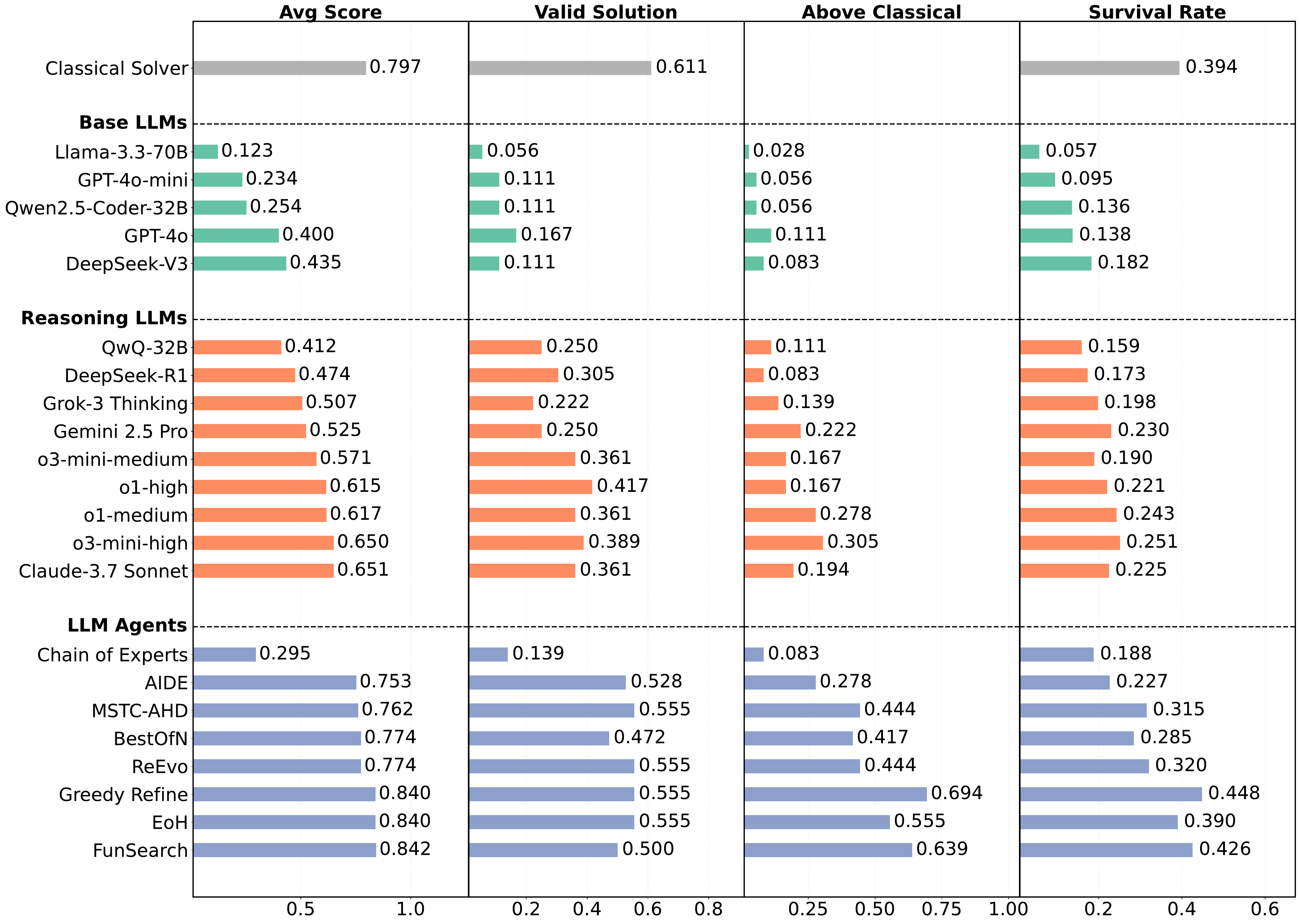}
    \caption{\textbf{Overall Performance.} 
    LLM Agents are all based on \texttt{o3-mini-medium.}
\textit{Avg Score} refers to the average normalized objective scores across all problems. 
\textit{Valid Solution} indicates the percentage of test-set problems for which the solutions are feasible.  
\textit{Above Classical} represents the percentage of test instances where the model outperforms the classical solver baseline. 
\textit{Survival Rate} measures the percentage of test instances where the model’s score exceeds 99\% of the reference score.}
    \label{fig:benchmark}
\end{figure*}
% \input{table/main}

% \paragraph{Human-designed CO Solvers}
% We evaluate two general CO solvers: \textit{Gurobi}~\citep{gurobi} and \textit{OR-Tools}~\citep{ortools}. To formulate the problems in the required format, we first use \textit{o3-mini-high} to draft the initial code, followed by manual revisions for correctness and completeness.
% \section{Experimental Results}

% See Appendix \ref{sec:agent-imp} for more implementation details.

\subsection{Implementation Details}
For benchmark evaluation, we limit the solving time of each test instance to 10 seconds on a single CPU, such that the exact solving of the problem (achieving the optimal solution) is impossible on most test instances. Test instances that result in a timeout or error receive a score of 0.

For agent implementation, we use o3-mini-medium as the default base model.
Since the original implementations of these agents may use different evaluation setups, we adapt their approaches to our benchmark setting (i.e., end-to-end algorithm search) by adjusting the prompts and tools.
For all agents, we set the number of iteration steps to $64$. In each step, the agent generates a code block as a candidate algorithm and obtains its evaluation score on the development set.
After 64 iterations, the agent produces 64 candidate algorithms, from which the best-performing solution on the development set is selected for final benchmark evaluation.
All evaluations are conducted on a single CPU core of a dual AMD EPYC 7313 16-Core processor.

\subsection{Main Results}
Figure~\ref{fig:benchmark} presents the results of LLMs and agents on the test set.
We highlight the following key findings.

\noindent\textbf{Direct generation performance is limited.} LLMs show significantly lower average scores compared to the classical solver. They often fail to generate valid solutions (i.e., bug-free code that satisfies all constraints within the time limit), rarely outperform the classical solver on individual instances, and often fail to produce optimal solutions. Reasoning-capable models tend to perform better than non-reasoning ones. The best-performing LLM for one-shot generation is \textit{Claude-3.7 Sonnet}, with an average score of 0.65.

\noindent\textbf{Agentic systems substantially improve LLM performance.}
Compared to direct generation, the agentic pipeline achieves considerably higher scores across all metrics. Among the evaluated frameworks, FunSearch attains the highest average score of 0.842, outperforming the classical solver (0.797). It also surpasses the solver on over half the test instances (see "Above Classical" score) and achieves a higher survival rate. These results highlight the effectiveness of LLM-based agents in solving CO problems.

\begin{figure*}[!ht]
    \centering
\includegraphics[width=0.9\linewidth]{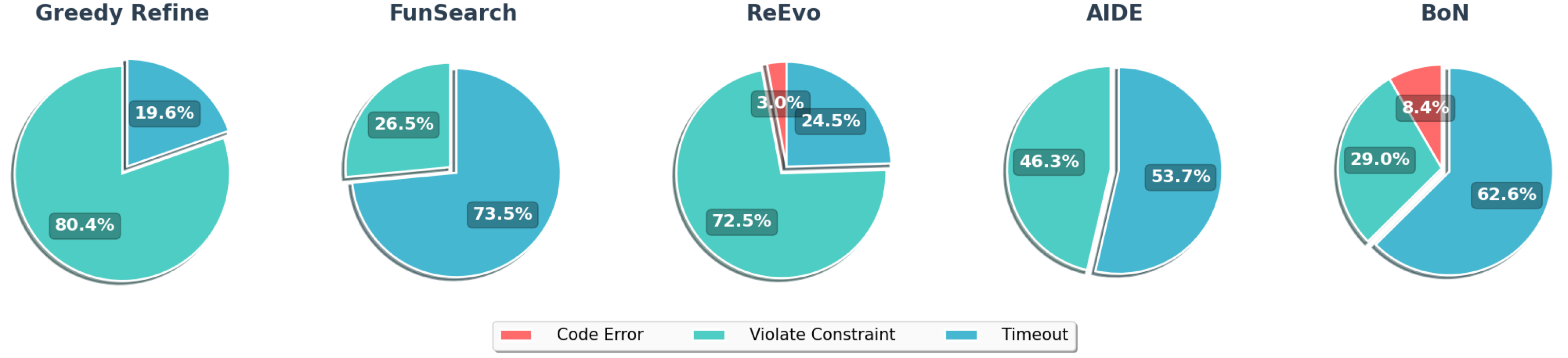}
    \caption{\textbf{Agents Error Analysis.} Distribution of three types of errors among invalid solutions for five agents.}
    \label{fig:pie}
\end{figure*}

\begin{figure}[t!]
  \centering
  \includegraphics[width=\linewidth]{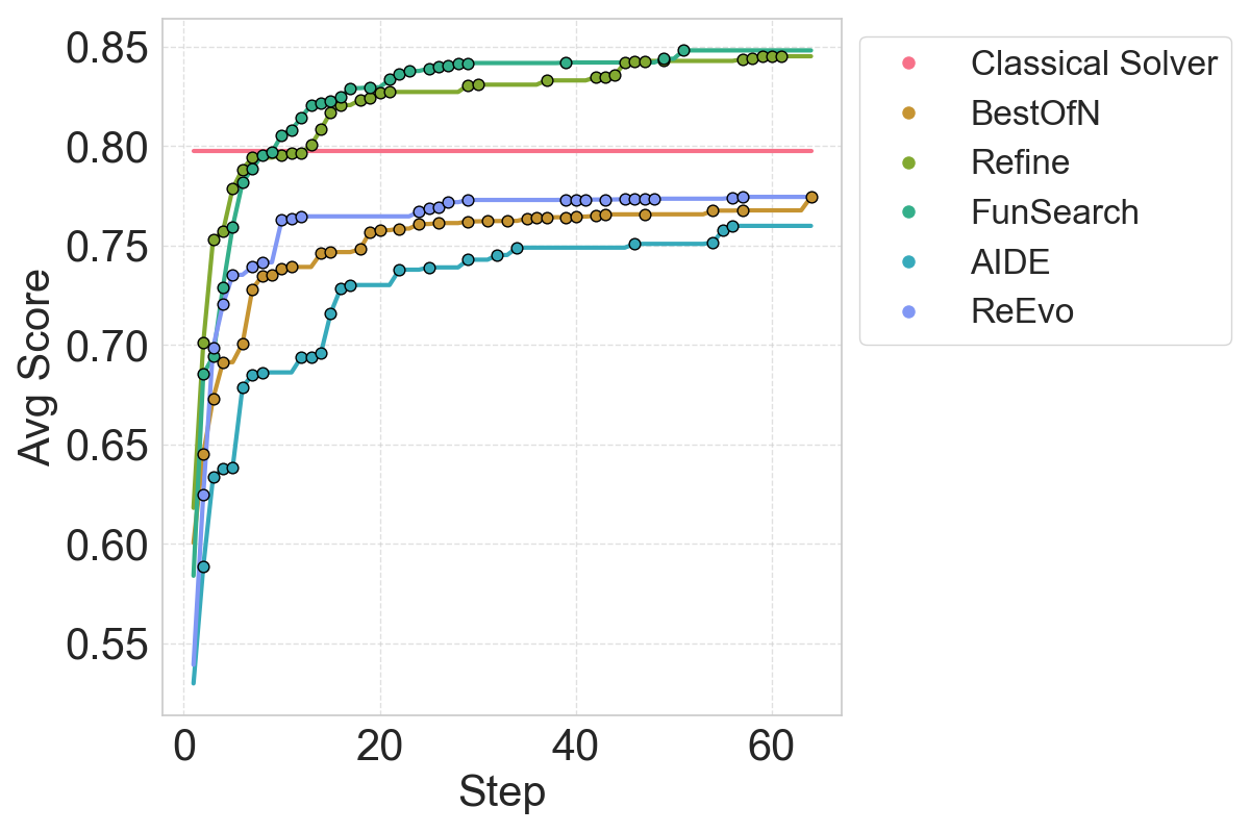}
  % \vspace*{-3mm}
  \caption{\textbf{Avg Score} vs. the number of iteration steps (in total 64 steps) during the algorithm development.}
  \label{fig:scaling}
\end{figure}

\begin{figure}
    \centering
    \includegraphics[width=\linewidth]{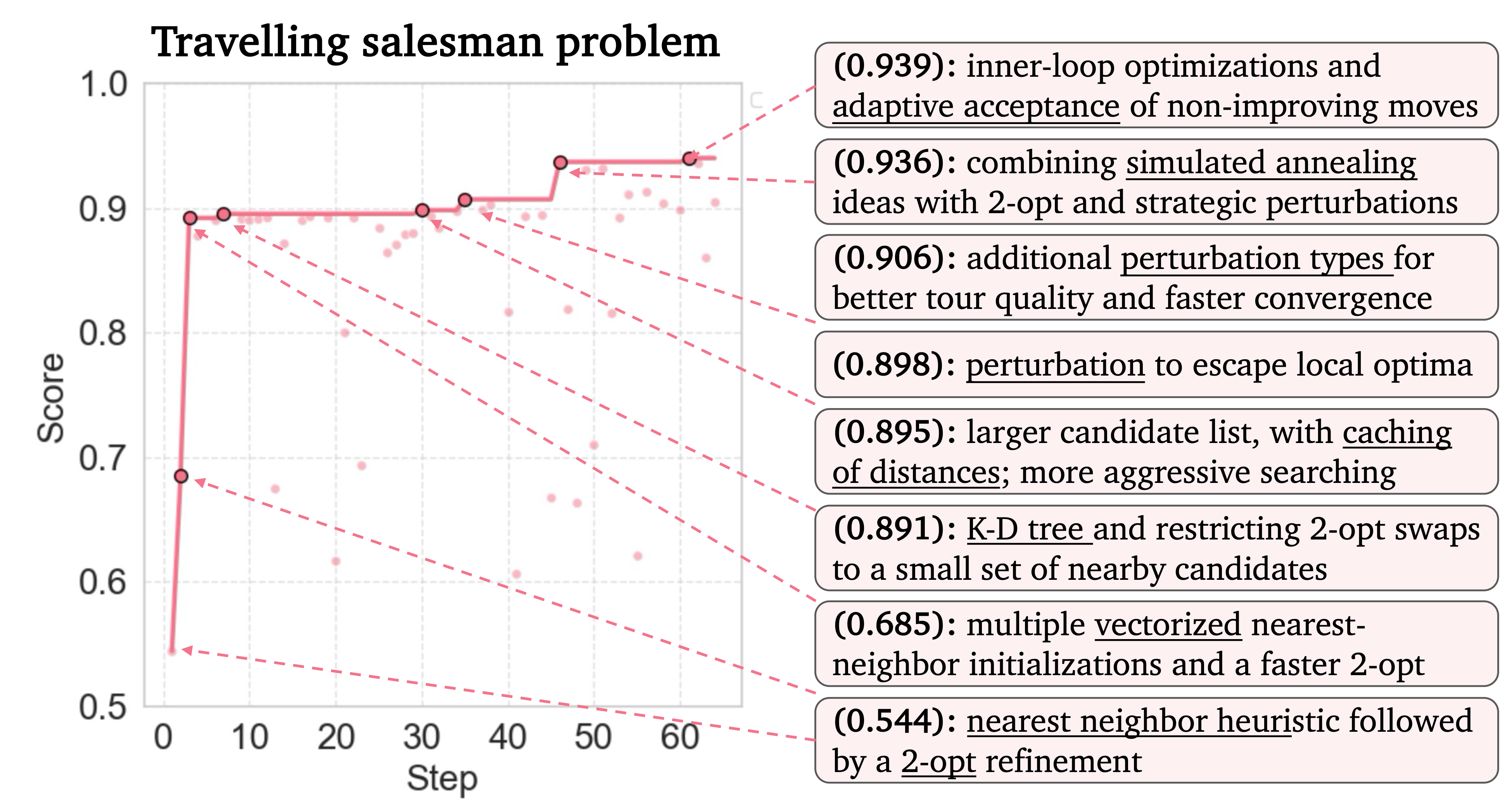}
    \caption{Trajectory of algorithm development for Greedy Refinement on TSP over 64 steps. The curve and highlighted dots indicate the best-ever score and the steps where improvements occurred. The algorithmic ideas behind each improvement step are summarized in corresponding boxes.}\label{fig:case}
\end{figure}

\noindent\textbf{Agent performance varies widely.} Some advanced agentic frameworks, such as AIDE, underperform compared to simpler strategies like BestOfN on most metrics, though they show higher valid solution rates—possibly due to their debugging capabilities. This indicates that current planning mechanisms in agents are still underdeveloped and may not reliably outperform random sampling.

\noindent\textbf{Valid solution rates still lag behind classical solvers.}
According to the \textit{Valid Solution} metric, the best-performing agents achieve a success rate of 0.555—lower than that of the classical solver (0.611). This suggests that current agents often struggle with solution feasibility and reliability.

\subsection{Agents Error Analysis}
To investigate why the agents' valid solution scores are low, Figure~\ref{fig:pie} shows the types of errors among invalid solutions for five agents. We observe that code errors (i.e., bugs that prevent compilation) are the least frequent issue. The dominant error type varies across agents: Greedy Refine and ReEvo exhibit more constraint violations, while FunSearch, AIDE, and BoN encounter more timeout errors.
This highlights agents’ limitations in satisfying constraints and generating efficient algorithms within time limits.

\subsection{Performance over Iteration Steps}
Figure~\ref{fig:scaling} illustrates the performance of several representative LLM agents across different iteration steps. At each step, the agent generates a new algorithm and receives evaluation results on the development set. We also include the performance of the classical solver baseline for comparison.

All agents exhibit the ability to improve their performance with more iteration steps. FunSearch consistently achieves the best results, reaching a score of 0.8423 and converging after around 50 steps. Notably, both FunSearch and Refine discover algorithms that outperform the classical solver within approximately 10 steps. However, performance tends to saturate after 30 steps, with further search yielding diminishing returns. Enabling more consistent improvements under longer search budgets presents an interesting future direction.

Figure~\ref{fig:case} shows an example trajectory of algorithm development by Greedy Refinement (o3-mini) on TSP over multiple search steps. In the early stages, the agent enhances code efficiency by adopting vectorized data structures and utilizing a K-D tree. It then increases the number of search iterations and introduces perturbations to escape local optima. Finally, the agent integrates simulated annealing to balance exploration and exploitation and applies adaptive heuristics for different instance sizes. This example demonstrates that LLMs excel in applying established techniques to improve efficiency and implementation quality, but failing at algorithmic novelty.

% We also observe that, reasoning models like o3-mini exhibit superior inference-scaling performance compared to non-reasoning models like GPT-4o. 

\begin{table*}[t!]
\centering
\setlength{\tabcolsep}{3pt}
\begin{tabular}{l | cccccc|cccc}
\toprule
& \multicolumn{2}{c}{TSP-500} & \multicolumn{2}{c}{TSP-1000} & \multicolumn{2}{c|}{TSP-10000} & \multicolumn{2}{c}{ER-Small} & \multicolumn{2}{c}{ER-Large} \\
% \hline
& Len $\downarrow$ & Time $\downarrow$ & Len $\downarrow$ & Time $\downarrow$ & Len $\downarrow$ & Time $\downarrow$ 
& Size $\uparrow$ & Time $\downarrow$ & Size $\uparrow$ & Time $\downarrow$\\
\midrule
Gurobi & 16.55 & 45.6h & - & - & - & - & 41.38 & 50.0m & - & - \\
DIMES  & 18.84 & 1.1m & 26.36 & 2.4m & 85.75 & 4.8m & 42.06 & 12.0m & 332.80 & 12.5m \\
% Diffusco (SL+S)  & 17.23 & 11.0m & 25.19 & 46.0m & 95.52 & 6.59h & 41.12 & 26.6m & - & - \\
DIFUSCO  & 16.65 & 11.5m & 23.45 & 48.1m & 73.89 & 6.72h & 41.12 & 26.6m & - & - \\
T2T & 16.61 & 16.0m & 23.30 & 54.6m & - & - & 41.37 & 29.7m & - & -\\
LEHD + ReEvo & 16.78 & - & 23.82 & - & - &  - & - & - & - & - \\
\midrule
Greedy Refine (o3-mini) & 17.37 & 19.1m & 24.40 & 19.1m & 77.65 & 2.5m & 42.35 & 20.1m & 354.00 & 2.5m\\
FunSearch (o3-mini) & 17.20 & 19.1m & 25.31 & 19.1m & 80.18 & 2.5m & 41.65 & 1.9m & 356.50 & 2.1m \\

\bottomrule
\end{tabular}
\caption{
Objective values and solving time of different solvers on TSP and MIS, with varying data sizes.
% Results of baselines are taken from the published results in corresponding papers.
}\label{table:tsp}
% \vspace*{-2mm}
\end{table*}

\subsection{Comparison to Neural Solvers}
Table \ref{table:tsp} compares the performance of agents with representative neural solvers on TSP and MIS, two well-studied CO problems. 
We include DIMES~\citep{qiu2022dimes}, DIFUSCO~\citep{Sun2023DIFUSCOGD}, and T2T~\citep{Li2023FromDL} as neural baselines. For the method with multiple variants, we only include their best results on each dataset.
We also consider a hybrid method, LEHD + ReEvo~\citep{ye2024reevo}, which combines the neural solver with LLM-designed heuristics.
We report both the objective values (the tour length for TSP and set size for MIS) and the solving time.
The results show that the agents such as Greedy Refine and FunSearch achieve competitive performance on both problems, often outperforming existing neural solvers under similar time budget and approaching the best results achieved by previous solvers given extended search time.
% We also observe that Greedy Refine performs slightly better than FunSearch on the two problems.

\begin{figure}
    \centering
    \includegraphics[width=0.9\linewidth]{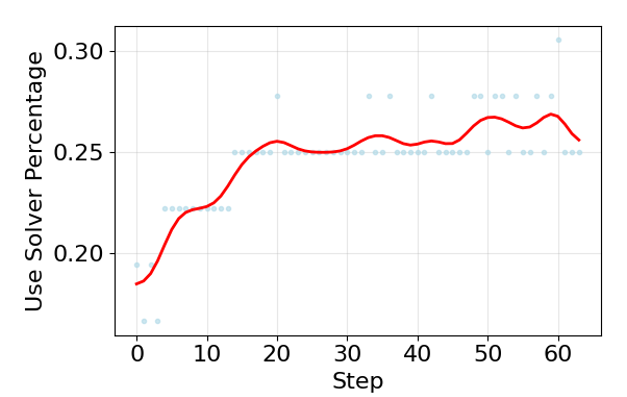}
    \caption{Percentage of algorithms developed by the Greedy Refinement agent that rely on existing solvers (e.g., code importing \texttt{ortools}, \texttt{pulp}) over 64 iteration steps. We observe an increasing use of existing solvers.}\label{fig:sol}
\end{figure}

\subsection{Solution Analysis}
In Figure~\ref{fig:sol}, we plot the percentage of algorithms developed by the Greedy Refinement agent for the 36 CO problems that utilize existing solvers (e.g., code importing \texttt{ortools}, \texttt{pulp}). The percentages are shown across 64 iteration steps. We observe an increasing trend in the use of existing solvers in the agent’s solutions. After 64 iterations, the final usage rate reaches 25\% (i.e., solutions for 9 problems use existing solvers). The solvers used throughout all steps and problems are limited to three: \texttt{ortools}, \texttt{pulp}, and \texttt{scipy}.

This suggests that while existing LLM agents are capable of developing algorithms without relying on existing solvers for most problems, there is a growing tendency to do so over time. Moreover, the solvers used are basic general-purpose tools rather than state-of-the-art solvers specifically designed for each problem (e.g., LKH for TSP), indicating that the agent lacks the necessary knowledge to select the best-performing solver.

% \begin{figure}[t!]
%   \centering
%   \includegraphics[width=.95\textwidth]{figs/case.png}
%   \caption{Trajectory of algorithm development for Greedy Refinement (o3-mini) on the traveling salesman problem (TSP) over 64 research steps. Each dot represents the evaluation score at a given step. The curve and highlighted dots indicate the best-ever score and the steps where improvements occurred. The algorithmic ideas behind each improvement step are summarized in the corresponding boxes.}\label{fig:case}
% \end{figure}

% It then explores a combination of simulated annealing and 2-Opt search, while also refining the heuristics used in priority functions.
% Finally, the agent develops a compute-adaptive algorithm that adjusts the pipeline based on input data scale, aiming to find the best solution within a preset time limit.

\section{Related Work}
\subsection{Automatic Algorithm Search for CO}
Automating algorithm search for combinatorial optimization (CO) has emerged as a significant research direction in the machine learning community. Traditional machine learning solvers primarily parameterize CO algorithms as trainable neural networks \citep{bengio2020machine, JMLR:v24:21-0449}. Although effective in capturing data distributions, these neural approaches often struggle to generate feasible solutions, necessitating integration with human-designed heuristics such as branch-and-bound \citep{conf/nips/GasseCFCL19} and tree search \citep{bother2022whats}. To address this limitation, \citet{kuang2024rethinking, pmlr-v235-kuang24a} propose to decompose CO algorithms into symbolic operators and conduct searches in the symbolic space. However, designing these unit symbolic operators demands substantial human expertise, limiting generalizability and comprehensive coverage of all algorithm types. Recent advances in Large Language Models (LLMs) and LLM-based agents have significantly mitigated this challenge by enabling symbolic searching in programming language formats \citep{RomeraParedes2023MathematicalDF, ye2024reevo, Liu2024EoH}. Building on these developments, CO-Bench aims to extend the success of these methods to more real-world CO problems and facilitate further research in this domain.

\subsection{CO Benchmarks for LLMs}
Existing CO benchmarks can be roughly classified into two categories. The first type  formulates CO problems as question-answering tasks \citep{fan-etal-2024-nphardeval, tang2025grapharena}. Although LLMs have the potential to solve CO problems via natural language reasoning, their excessive parameter size makes them inefficient CO solvers in general. Therefore, the second type of benchmarks  evaluates the tool-using ability of LLMs, e.g., calling an existing CO solver, to address CO problems \citep{xiao2024chainofexperts, pmlr-v235-ahmaditeshnizi24a, yang2025optibench}. However, these benchmarks only evaluate the correctness of the generated algorithm on small-scale CO problems, whose problem parameters could be fully expressed in natural language. In contrast, CO-Bench targets scientific and industrial challenges, emphasizing the evaluation of algorithm efficiency on diverse, large-scale CO instances. This results in a more demanding benchmark, well-suited for assessing powerful reasoning models and  agents.

% \subsection{Automatic Heuristic Discovery for Combinatorial Optimization}

\section{Conclusion}
This work introduces CO-Bench, the first benchmark designed to evaluate the ability of LLMs in the search of combinatorial optimization (CO) algorithms. Our systematic evaluation reveals that reasoning-focused LLMs, especially when paired with agentic frameworks, can automatically discover effective algorithms that rival or surpass the classical solvers designed by human experts, with competitive searching time. However, we also identify key limitations of current LLM agents such as they struggle to understand the problem constraints. These shortcomings highlight the need for future research to enhance agents’ problem comprehension and creative reasoning abilities in CO tasks, enabling more robust and autonomous scientific discovery.

% \subsection{Impact of Reasoning Length on Performance}

\bibliography{aaai2026,orlib}

\begin{thebibliography}{85}
\providecommand{\natexlab}[1]{#1}

\bibitem[{Ahmaditeshnizi, Gao, and Udell(2024)}]{pmlr-v235-ahmaditeshnizi24a}
Ahmaditeshnizi, A.; Gao, W.; and Udell, M. 2024.
\newblock {O}pti{MUS}: Scalable Optimization Modeling with ({MI}){LP} Solvers and Large Language Models.
\newblock In Salakhutdinov, R.; Kolter, Z.; Heller, K.; Weller, A.; Oliver, N.; Scarlett, J.; and Berkenkamp, F., eds., \emph{Proceedings of the 41st International Conference on Machine Learning}, volume 235 of \emph{Proceedings of Machine Learning Research}, 577--596. PMLR.

\bibitem[{and(1990)}]{Beasley01021990}
and, J. E.~B. 1990.
\newblock Linear Programming on Cray Supercomputers.
\newblock \emph{Journal of the Operational Research Society}, 41(2): 133--139.

\bibitem[{Anken and Beasley(2012)}]{Anken2012CorporateSO}
Anken, F.; and Beasley, J.~E. 2012.
\newblock Corporate structure optimisation for multinational companies.
\newblock \emph{Omega-international Journal of Management Science}, 40: 230--243.

\bibitem[{Anthropic(2025)}]{claude_sonnet}
Anthropic. 2025.
\newblock Claude Sonnet.
\newblock \url{https://www.anthropic.com/claude/sonnet}.
\newblock Accessed: 2025-03-24.

\bibitem[{Beasley(1985{\natexlab{a}})}]{Beasley1985AnAF}
Beasley, J.~E. 1985{\natexlab{a}}.
\newblock An algorithm for the two-dimensional assortment problem.
\newblock \emph{European Journal of Operational Research}, 19: 253--261.

\bibitem[{Beasley(1985{\natexlab{b}})}]{Beasley1985AlgorithmsFU}
Beasley, J.~E. 1985{\natexlab{b}}.
\newblock Algorithms for Unconstrained Two-Dimensional Guillotine Cutting.
\newblock \emph{Journal of the Operational Research Society}, 36: 297--306.

\bibitem[{Beasley(1985{\natexlab{c}})}]{Beasley1985AnET}
Beasley, J.~E. 1985{\natexlab{c}}.
\newblock An Exact Two-Dimensional Non-Guillotine Cutting Tree Search Procedure.
\newblock \emph{Oper. Res.}, 33: 49--64.

\bibitem[{Beasley(1985{\natexlab{d}})}]{Beasley1985ANO}
Beasley, J.~E. 1985{\natexlab{d}}.
\newblock A note on solving large p-median problems.
\newblock \emph{European Journal of Operational Research}, 21: 270--273.

\bibitem[{Beasley(1988)}]{Beasley1988AnAF}
Beasley, J.~E. 1988.
\newblock An algorithm for solving large capacitated warehouse location problems.
\newblock \emph{European Journal of Operational Research}, 33: 314--325.

\bibitem[{Beasley(1990)}]{Beasley1990ORLibraryDT}
Beasley, J.~E. 1990.
\newblock OR-Library: Distributing Test Problems by Electronic Mail.
\newblock \emph{Journal of the Operational Research Society}, 41: 1069--1072.

\bibitem[{Beasley(1992)}]{Beasley1992AHF}
Beasley, J.~E. 1992.
\newblock A heuristic for Euclidean and rectilinear Steiner problems.
\newblock \emph{European Journal of Operational Research}, 58: 284--292.

\bibitem[{Beasley(1993)}]{Beasley1993LagrangeanHF}
Beasley, J.~E. 1993.
\newblock Lagrangean heuristics for location problems.
\newblock \emph{European Journal of Operational Research}, 65: 383--399.

\bibitem[{Beasley(2004)}]{Beasley2004APH}
Beasley, J.~E. 2004.
\newblock A population heuristic for constrained two-dimensional non-guillotine cutting.
\newblock \emph{Eur. J. Oper. Res.}, 156: 601--627.

\bibitem[{Beasley and Cao(1996)}]{Beasley1996ATS}
Beasley, J.~E.; and Cao, B. 1996.
\newblock A tree search algorithm for the crew scheduling problem.
\newblock \emph{European Journal of Operational Research}, 94: 517--526.

\bibitem[{Beasley and Christofides(1989)}]{Beasley1989AnAF}
Beasley, J.~E.; and Christofides, N. 1989.
\newblock An algorithm for the resource constrained shortest path problem.
\newblock \emph{Networks}, 19: 379--394.

\bibitem[{Beasley and J{\"o}rnsten(1992)}]{Beasley1992EnhancingAA}
Beasley, J.~E.; and J{\"o}rnsten, K. 1992.
\newblock Enhancing an algorithm for set covering problems.
\newblock \emph{European Journal of Operational Research}, 58: 293--300.

\bibitem[{Beasley et~al.(2000)Beasley, Krishnamoorthy, Sharaiha, and Abramson}]{Beasley2000SchedulingAL}
Beasley, J.~E.; Krishnamoorthy, M.; Sharaiha, Y.~M.; and Abramson, D. 2000.
\newblock Scheduling Aircraft Landings - The Static Case.
\newblock \emph{Transp. Sci.}, 34: 180--197.

\bibitem[{Beasley et~al.(2004)Beasley, Krishnamoorthy, Sharaiha, and Abramson}]{Beasley2004DisplacementPA}
Beasley, J.~E.; Krishnamoorthy, M.; Sharaiha, Y.~M.; and Abramson, D. 2004.
\newblock Displacement problem and dynamically scheduling aircraft landings.
\newblock \emph{Journal of the Operational Research Society}, 55: 54--64.

\bibitem[{Bengio, Lodi, and Prouvost(2020)}]{bengio2020machine}
Bengio, Y.; Lodi, A.; and Prouvost, A. 2020.
\newblock Machine Learning for Combinatorial Optimization: a Methodological Tour d'Horizon.
\newblock arXiv:1811.06128.

\bibitem[{Berthold(2006)}]{Berthold2006PrimalHF}
Berthold, T. 2006.
\newblock \emph{Primal heuristics for mixed integer programs}.
\newblock Ph.D. thesis, Zuse Institute Berlin (ZIB).

\bibitem[{Bischoff(2006)}]{Bischoff2006ThreedimensionalPO}
Bischoff, E.~E. 2006.
\newblock Three-dimensional packing of items with limited load bearing strength.
\newblock \emph{Eur. J. Oper. Res.}, 168: 952--966.

\bibitem[{Bischoff and Ratcliff(1995)}]{Bischoff1995IssuesIT}
Bischoff, E.~E.; and Ratcliff, M. S.~W. 1995.
\newblock Issues in the development of approaches to container loading.
\newblock \emph{Omega-international Journal of Management Science}, 23: 377--390.

\bibitem[{Biskup and Feldmann(2001)}]{Biskup2001BenchmarksFS}
Biskup, D.; and Feldmann, M. 2001.
\newblock Benchmarks for scheduling on a single machine against restrictive and unrestrictive common due dates.
\newblock \emph{Comput. Oper. Res.}, 28: 787--801.

\bibitem[{B{\"o}ther et~al.(2022)B{\"o}ther, Ki{\ss}ig, Taraz, Cohen, Seidel, and Friedrich}]{bother2022whats}
B{\"o}ther, M.; Ki{\ss}ig, O.; Taraz, M.; Cohen, S.; Seidel, K.; and Friedrich, T. 2022.
\newblock What{\textquoteright}s Wrong with Deep Learning in Tree Search for Combinatorial Optimization.
\newblock In \emph{International Conference on Learning Representations}.

\bibitem[{Cappanera and Trubian(2001)}]{Cappanera2001ALH}
Cappanera, P.; and Trubian, M. 2001.
\newblock A Local-Search-Based Heuristic for the Demand-Constrained Multidimensional Knapsack Problem.
\newblock \emph{INFORMS J. Comput.}, 17: 82--98.

\bibitem[{Cappart et~al.(2023)Cappart, ChÃ©telat, Khalil, Lodi, Morris, and VeliÄkoviÄ‡}]{JMLR:v24:21-0449}
Cappart, Q.; ChÃ©telat, D.; Khalil, E.~B.; Lodi, A.; Morris, C.; and VeliÄkoviÄ‡, P. 2023.
\newblock Combinatorial Optimization and Reasoning with Graph Neural Networks.
\newblock \emph{Journal of Machine Learning Research}, 24(130): 1--61.

\bibitem[{Chakhlevitch and Glass(2009)}]{Chakhlevitch2009SchedulingRJ}
Chakhlevitch, K.; and Glass, C.~A. 2009.
\newblock Scheduling reentrant jobs on parallel machines with a remote server.
\newblock \emph{Comput. Oper. Res.}, 36: 2580--2589.

\bibitem[{Chan et~al.(2024)Chan, Chowdhury, Jaffe, Aung, Sherburn, Mays, Starace, Liu, Maksin, Patwardhan, Weng, and Mkadry}]{Chan2024MLEbenchEM}
Chan, J.~S.; Chowdhury, N.; Jaffe, O.; Aung, J.; Sherburn, D.; Mays, E.; Starace, G.; Liu, K.; Maksin, L.; Patwardhan, T.~A.; Weng, L.; and Mkadry, A. 2024.
\newblock MLE-bench: Evaluating Machine Learning Agents on Machine Learning Engineering.
\newblock \emph{ArXiv}, abs/2410.07095.

\bibitem[{Chen et~al.(2021)Chen, Tworek, Jun, Yuan, Pond{\'e}, Kaplan, Edwards, Burda, Joseph, Brockman, Ray, Puri, Krueger, Petrov, Khlaaf, Sastry, Mishkin, Chan, Gray, Ryder, Pavlov, Power, Kaiser, Bavarian, Winter, Tillet, Such, Cummings, Plappert, Chantzis, Barnes, Herbert-Voss, Guss, Nichol, Babuschkin, Balaji, Jain, Carr, Leike, Achiam, Misra, Morikawa, Radford, Knight, Brundage, Murati, Mayer, Welinder, McGrew, Amodei, McCandlish, Sutskever, and Zaremba}]{Chen2021EvaluatingLL}
Chen, M.; Tworek, J.; Jun, H.; Yuan, Q.; Pond{\'e}, H.; Kaplan, J.; Edwards, H.; Burda, Y.; Joseph, N.; Brockman, G.; Ray, A.; Puri, R.; Krueger, G.; Petrov, M.; Khlaaf, H.; Sastry, G.; Mishkin, P.; Chan, B.; Gray, S.; Ryder, N.; Pavlov, M.; Power, A.; Kaiser, L.; Bavarian, M.; Winter, C.; Tillet, P.; Such, F.~P.; Cummings, D.~W.; Plappert, M.; Chantzis, F.; Barnes, E.; Herbert-Voss, A.; Guss, W.~H.; Nichol, A.; Babuschkin, I.; Balaji, S.; Jain, S.; Carr, A.; Leike, J.; Achiam, J.; Misra, V.; Morikawa, E.; Radford, A.; Knight, M.~M.; Brundage, M.; Murati, M.; Mayer, K.; Welinder, P.; McGrew, B.; Amodei, D.; McCandlish, S.; Sutskever, I.; and Zaremba, W. 2021.
\newblock Evaluating Large Language Models Trained on Code.
\newblock \emph{ArXiv}, abs/2107.03374.

\bibitem[{Christofides and Beasley(1984)}]{Christofides1984ThePR}
Christofides, N.; and Beasley, J.~E. 1984.
\newblock The period routing problem.
\newblock \emph{Networks}, 14: 237--256.

\bibitem[{Christofides and Whitlock(1977)}]{Christofides1977AnAF}
Christofides, N.; and Whitlock, C. 1977.
\newblock An Algorithm for Two-Dimensional Cutting Problems.
\newblock \emph{Oper. Res.}, 25: 30--44.

\bibitem[{Chu and Beasley(1998)}]{Chu1998ConstraintHI}
Chu, P.~C.; and Beasley, J.~E. 1998.
\newblock Constraint Handling in Genetic Algorithms: The Set Partitioning Problem.
\newblock \emph{Journal of Heuristics}, 4: 323--357.

\bibitem[{Crama(1997)}]{CRAMA1997136}
Crama, Y. 1997.
\newblock Combinatorial optimization models for production scheduling in automated manufacturing systems.
\newblock \emph{European Journal of Operational Research}, 99(1): 136--153.

\bibitem[{DeepMind(2025)}]{gemini_flash_thinking}
DeepMind, G. 2025.
\newblock Flash Thinking: Behind the Scenes of Gemini.
\newblock \url{https://deepmind.google/technologies/gemini/flash-thinking/}.
\newblock Accessed: 2025-03-24.

\bibitem[{DeepSeek-AI(2024)}]{DeepSeekAI2024DeepSeekV3TR}
DeepSeek-AI. 2024.
\newblock DeepSeek-V3 Technical Report.
\newblock \emph{ArXiv}, abs/2412.19437.

\bibitem[{DeepSeek-AI(2025{\natexlab{a}})}]{deepseekai2025deepseekr1incentivizingreasoningcapability}
DeepSeek-AI. 2025{\natexlab{a}}.
\newblock DeepSeek-R1: Incentivizing Reasoning Capability in LLMs via Reinforcement Learning.
\newblock arXiv:2501.12948.

\bibitem[{DeepSeek-AI(2025{\natexlab{b}})}]{DeepSeekAI2025DeepSeekR1IR}
DeepSeek-AI. 2025{\natexlab{b}}.
\newblock DeepSeek-R1: Incentivizing Reasoning Capability in LLMs via Reinforcement Learning.
\newblock \emph{ArXiv}, abs/2501.12948.

\bibitem[{Erdos and R{\'e}nyi(1984)}]{Erdos1984OnTE}
Erdos, P.~L.; and R{\'e}nyi, A. 1984.
\newblock On the evolution of random graphs.
\newblock \emph{Transactions of the American Mathematical Society}, 286: 257--257.

\bibitem[{Falkenauer(1996)}]{Falkenauer1996AHG}
Falkenauer, E. 1996.
\newblock A hybrid grouping genetic algorithm for bin packing.
\newblock \emph{Journal of Heuristics}, 2: 5--30.

\bibitem[{Fan et~al.(2024)Fan, Hua, Li, Ling, and Zhang}]{fan-etal-2024-nphardeval}
Fan, L.; Hua, W.; Li, L.; Ling, H.; and Zhang, Y. 2024.
\newblock {NPH}ard{E}val: Dynamic Benchmark on Reasoning Ability of Large Language Models via Complexity Classes.
\newblock In Ku, L.-W.; Martins, A.; and Srikumar, V., eds., \emph{Proceedings of the 62nd Annual Meeting of the Association for Computational Linguistics (Volume 1: Long Papers)}, 4092--4114. Bangkok, Thailand: Association for Computational Linguistics.

\bibitem[{Fleurent and Ferland(1996)}]{Fleurent1996GeneticAH}
Fleurent, C.; and Ferland, J.~A. 1996.
\newblock Genetic and hybrid algorithms for graph coloring.
\newblock \emph{Annals of Operations Research}, 63: 437--461.

\bibitem[{Gasse et~al.(2019)Gasse, Chételat, Ferroni, Charlin, and Lodi}]{conf/nips/GasseCFCL19}
Gasse, M.; Chételat, D.; Ferroni, N.; Charlin, L.; and Lodi, A. 2019.
\newblock Exact Combinatorial Optimization with Graph Convolutional Neural Networks.
\newblock In \emph{Advances in Neural Information Processing Systems 32}.

\bibitem[{Gottweis et~al.(2025)Gottweis, Weng, Daryin, Tu, Palepu, Sirkovic, Myaskovsky, Weissenberger, Rong, Tanno, Saab, Popovici, Blum, Zhang, Chou, Hassidim, Gokturk, Vahdat, Kohli, Matias, Carroll, Kulkarni, Tomaev, Guan, Dhillon, Vaishnav, Lee, Costa, Penad'es, Peltz, Xu, Pawlosky, Karthikesalingam, and Natarajan}]{Gottweis2025TowardsAA}
Gottweis, J.; Weng, W.-H.; Daryin, A.; Tu, T.; Palepu, A.; Sirkovic, P.; Myaskovsky, A.; Weissenberger, F.; Rong, K.; Tanno, R.; Saab, K.; Popovici, D.; Blum, J.; Zhang, F.; Chou, K.; Hassidim, A.; Gokturk, B.; Vahdat, A.; Kohli, P.; Matias, Y.; Carroll, A.; Kulkarni, K.; Tomaev, N.; Guan, Y.; Dhillon, V.; Vaishnav, E.~D.; Lee, B.; Costa, T. R.~D.; Penad'es, J.~R.; Peltz, G.; Xu, Y.; Pawlosky, A.; Karthikesalingam, A.; and Natarajan, V. 2025.
\newblock Towards an AI co-scientist.
\newblock \emph{ArXiv}, abs/2502.18864.

\bibitem[{Gusfield(1997)}]{gusfield1997algorithms}
Gusfield, D. 1997.
\newblock Algorithms on stings, trees, and sequences: Computer science and computational biology.
\newblock \emph{Acm Sigact News}, 28(4): 41--60.

\bibitem[{Hui et~al.(2024)Hui, Yang, Cui, Yang, Liu, Zhang, Liu, Zhang, Yu, Dang, Yang, Men, Huang, Quan, Ren, Ren, Zhou, and Lin}]{Hui2024Qwen25CoderTR}
Hui, B.; Yang, J.; Cui, Z.; Yang, J.; Liu, D.; Zhang, L.; Liu, T.; Zhang, J.; Yu, B.; Dang, K.; Yang, A.; Men, R.; Huang, F.; Quan, S.; Ren, X.; Ren, X.; Zhou, J.; and Lin, J. 2024.
\newblock Qwen2.5-Coder Technical Report.
\newblock \emph{ArXiv}, abs/2409.12186.

\bibitem[{Ivancic(1988)}]{Ivancic1988AnIP}
Ivancic, N.~J. 1988.
\newblock An integer programming based heuristic approach to the three dimensional packing problem.

\bibitem[{Jiang et~al.(2025)Jiang, Schmidt, Srikanth, Xu, Kaplan, Jacenko, and Wu}]{Jiang2025AIDEAE}
Jiang, Z.; Schmidt, D.; Srikanth, D.; Xu, D.; Kaplan, I.; Jacenko, D.; and Wu, Y. 2025.
\newblock AIDE: AI-Driven Exploration in the Space of Code.
\newblock \emph{ArXiv}, abs/2502.13138.

\bibitem[{Jimenez et~al.(2023)Jimenez, Yang, Wettig, Yao, Pei, Press, and Narasimhan}]{Jimenez2023SWEbenchCL}
Jimenez, C.~E.; Yang, J.; Wettig, A.; Yao, S.; Pei, K.; Press, O.; and Narasimhan, K. 2023.
\newblock SWE-bench: Can Language Models Resolve Real-World GitHub Issues?
\newblock \emph{ArXiv}, abs/2310.06770.

\bibitem[{Kuang et~al.(2024{\natexlab{a}})Kuang, Wang, Liu, Zhu, Li, Zeng, HAO, Li, and Wu}]{kuang2024rethinking}
Kuang, Y.; Wang, J.; Liu, H.; Zhu, F.; Li, X.; Zeng, J.; HAO, J.; Li, B.; and Wu, F. 2024{\natexlab{a}}.
\newblock Rethinking Branching on Exact Combinatorial Optimization Solver: The First Deep Symbolic Discovery Framework.
\newblock In \emph{The Twelfth International Conference on Learning Representations}.

\bibitem[{Kuang et~al.(2024{\natexlab{b}})Kuang, Wang, Zhou, Li, Zhu, Hao, and Wu}]{pmlr-v235-kuang24a}
Kuang, Y.; Wang, J.; Zhou, Y.; Li, X.; Zhu, F.; Hao, J.; and Wu, F. 2024{\natexlab{b}}.
\newblock Towards General Algorithm Discovery for Combinatorial Optimization: Learning Symbolic Branching Policy from Bipartite Graph.
\newblock In Salakhutdinov, R.; Kolter, Z.; Heller, K.; Weller, A.; Oliver, N.; Scarlett, J.; and Berkenkamp, F., eds., \emph{Proceedings of the 41st International Conference on Machine Learning}, volume 235 of \emph{Proceedings of Machine Learning Research}, 25623--25641. PMLR.

\bibitem[{Laporte(1992)}]{LAPORTE1992231}
Laporte, G. 1992.
\newblock The traveling salesman problem: An overview of exact and approximate algorithms.
\newblock \emph{European Journal of Operational Research}, 59(2): 231--247.

\bibitem[{Li et~al.(2023)Li, Guo, Wang, and Yan}]{Li2023FromDL}
Li, Y.; Guo, J.; Wang, R.; and Yan, J. 2023.
\newblock From Distribution Learning in Training to Gradient Search in Testing for Combinatorial Optimization.
\newblock In \emph{Neural Information Processing Systems}.

\bibitem[{Liu et~al.(2024)Liu, Tong, Yuan, Lin, Luo, Wang, Lu, and Zhang}]{Liu2024EoH}
Liu, F.; Tong, X.; Yuan, M.; Lin, X.; Luo, F.; Wang, Z.; Lu, Z.; and Zhang, Q. 2024.
\newblock Evolution of Heuristics: Towards Efficient Automatic Algorithm Design Using Large Language Model.
\newblock In \emph{ICML}.

\bibitem[{L{\'o}pez and Beasley(2016)}]{Lpez2016AFS}
L{\'o}pez, C.~O.; and Beasley, J.~E. 2016.
\newblock A formulation space search heuristic for packing unequal circles in a fixed size circular container.
\newblock \emph{Eur. J. Oper. Res.}, 251: 64--73.

\bibitem[{L{\'o}pez and Beasley(2018)}]{Lpez2018PackingUR}
L{\'o}pez, C.~O.; and Beasley, J.~E. 2018.
\newblock Packing unequal rectangles and squares in a fixed size circular container using formulation space search.
\newblock \emph{Comput. Oper. Res.}, 94: 106--117.

\bibitem[{Madaan et~al.(2023)Madaan, Tandon, Gupta, Hallinan, Gao, Wiegreffe, Alon, Dziri, Prabhumoye, Yang, Welleck, Majumder, Gupta, Yazdanbakhsh, and Clark}]{Madaan2023SelfRefineIR}
Madaan, A.; Tandon, N.; Gupta, P.; Hallinan, S.; Gao, L.; Wiegreffe, S.; Alon, U.; Dziri, N.; Prabhumoye, S.; Yang, Y.; Welleck, S.; Majumder, B.~P.; Gupta, S.; Yazdanbakhsh, A.; and Clark, P. 2023.
\newblock Self-Refine: Iterative Refinement with Self-Feedback.
\newblock \emph{ArXiv}, abs/2303.17651.

\bibitem[{Meta(2024)}]{Dubey2024TheL3}
Meta. 2024.
\newblock The Llama 3 Herd of Models.
\newblock \emph{ArXiv}, abs/2407.21783.

\bibitem[{Mingers and O'Brien(1995)}]{Mingers1995CreatingSG}
Mingers, J.~C.; and O'Brien, F.~A. 1995.
\newblock Creating student groups with similar characteristics: A heuristic approach.
\newblock \emph{Omega-international Journal of Management Science}, 23: 313--321.

\bibitem[{Motwani and Raghavan(2013)}]{Rajeev2013RA}
Motwani, R.; and Raghavan, P. 2013.
\newblock \emph{Randomized Algorithms}.
\newblock USA: Cambridge University Press.
\newblock ISBN 0511814070.

\bibitem[{Novikov et~al.(2025)Novikov, V˜u, Eisenberger, Dupont, Huang, Wagner, Shirobokov, Kozlovskii, Ruiz, Mehrabian, Kumar, See, Chaudhuri, Holland, Davies, Nowozin, Kohli, Balog, and Deepmind}]{Novikov2025AlphaEvolveAC}
Novikov, A.; V˜u, N.; Eisenberger, M.; Dupont, E.; Huang, P.-S.; Wagner, A.~Z.; Shirobokov, S.; Kozlovskii, B.~M.; Ruiz, F. J.~R.; Mehrabian, A.; Kumar, M.~P.; See, A.; Chaudhuri, S.; Holland, G.; Davies, A.; Nowozin, S.; Kohli, P.; Balog, M.; and Deepmind, G. 2025.
\newblock AlphaEvolve: A coding agent for scientific and algorithmic discovery.
\newblock \emph{ArXiv}, abs/2506.13131.

\bibitem[{OpenAI(2024{\natexlab{a}})}]{Hurst2024GPT4oSC}
OpenAI. 2024{\natexlab{a}}.
\newblock GPT-4o System Card.
\newblock \emph{ArXiv}, abs/2410.21276.

\bibitem[{OpenAI(2024{\natexlab{b}})}]{openai2024openaio1card}
OpenAI. 2024{\natexlab{b}}.
\newblock OpenAI o1 System Card.
\newblock arXiv:2412.16720.

\bibitem[{OpenAI(2025)}]{ZhangOpenAIOS}
OpenAI. 2025.
\newblock OpenAI o3-mini System Card.

\bibitem[{Osman(1995)}]{Osman1995HeuristicsFT}
Osman, I.~H. 1995.
\newblock Heuristics for the generalised assignment problem: simulated annealing and tabu search approaches.
\newblock \emph{Operations-Research-Spektrum}, 17: 211--225.

\bibitem[{Osman and Christofides(1994)}]{Osman1994CapacitatedCP}
Osman, I.~H.; and Christofides, N. 1994.
\newblock Capacitated clustering problems by hybrid simulated annealing and tabu search.
\newblock \emph{International Transactions in Operational Research}, 1: 317--336.

\bibitem[{Papadimitriou and Steiglitz(1982)}]{christos1982co}
Papadimitriou, C.; and Steiglitz, K. 1982.
\newblock \emph{Combinatorial Optimization: Algorithms and Complexity}, volume~32.
\newblock Courier Corporation.
\newblock ISBN 0-13-152462-3.

\bibitem[{Petersen(1967)}]{Petersen1967ComputationalEW}
Petersen, C.~C. 1967.
\newblock Computational Experience with Variants of the Balas Algorithm Applied to the Selection of R\&D Projects.
\newblock \emph{Management Science}, 13: 736--750.

\bibitem[{Qiu, Sun, and Yang(2022)}]{qiu2022dimes}
Qiu, R.; Sun, Z.; and Yang, Y. 2022.
\newblock {DIMES}: A Differentiable Meta Solver for Combinatorial Optimization Problems.
\newblock In Oh, A.~H.; Agarwal, A.; Belgrave, D.; and Cho, K., eds., \emph{Advances in Neural Information Processing Systems}.

\bibitem[{Qwen(2025)}]{qwq_32b}
Qwen. 2025.
\newblock QwQ-32B: Embracing the Power of Reinforcement Learning.
\newblock \url{https://qwenlm.github.io/blog/qwq-32b/}.
\newblock Accessed: 2025-03-24.

\bibitem[{Ramamonjison et~al.(2023)Ramamonjison, Yu, Li, Li, Carenini, Ghaddar, He, Mostajabdaveh, Banitalebi-Dehkordi, Zhou, and Zhang}]{Ramamonjison2023NL4OptCF}
Ramamonjison, R.; Yu, T.~T.; Li, R.; Li, H.; Carenini, G.; Ghaddar, B.; He, S.; Mostajabdaveh, M.; Banitalebi-Dehkordi, A.; Zhou, Z.; and Zhang, Y. 2023.
\newblock NL4Opt Competition: Formulating Optimization Problems Based on Their Natural Language Descriptions.
\newblock In \emph{Neural Information Processing Systems}.

\bibitem[{Ratcliff and Bischoff(1998)}]{Ratcliff1998AllowingFW}
Ratcliff, M. S.~W.; and Bischoff, E.~E. 1998.
\newblock Allowing for weight considerations in container loading.
\newblock \emph{Operations-Research-Spektrum}, 20: 65--71.

\bibitem[{Romera-Paredes et~al.(2023)Romera-Paredes, Barekatain, Novikov, Balog, Kumar, Dupont, Ruiz, Ellenberg, Wang, Fawzi, Kohli, Fawzi, Grochow, Lodi, Mouret, Ringer, and Yu}]{RomeraParedes2023MathematicalDF}
Romera-Paredes, B.; Barekatain, M.; Novikov, A.; Balog, M.; Kumar, M.~P.; Dupont, E.; Ruiz, F. J.~R.; Ellenberg, J.~S.; Wang, P.; Fawzi, O.; Kohli, P.; Fawzi, A.; Grochow, J.; Lodi, A.; Mouret, J.-B.; Ringer, T.; and Yu, T. 2023.
\newblock Mathematical discoveries from program search with large language models.
\newblock \emph{Nature}, 625: 468 -- 475.

\bibitem[{Shinn et~al.(2023)Shinn, Cassano, Labash, Gopinath, Narasimhan, and Yao}]{Shinn2023ReflexionLA}
Shinn, N.; Cassano, F.; Labash, B.; Gopinath, A.; Narasimhan, K.; and Yao, S. 2023.
\newblock Reflexion: language agents with verbal reinforcement learning.
\newblock In \emph{Neural Information Processing Systems}.

\bibitem[{Sun and Yang(2023)}]{Sun2023DIFUSCOGD}
Sun, Z.; and Yang, Y. 2023.
\newblock DIFUSCO: Graph-based Diffusion Solvers for Combinatorial Optimization.
\newblock \emph{ArXiv}, abs/2302.08224.

\bibitem[{Taillard(1993)}]{taillard1993benchmarks}
Taillard, E. 1993.
\newblock Benchmarks for basic scheduling problems.
\newblock \emph{European Journal of Operational Research}, 64(2): 278--285.

\bibitem[{Tang et~al.(2025)Tang, Zhang, Li, Chen, and Li}]{tang2025grapharena}
Tang, J.; Zhang, Q.; Li, Y.; Chen, N.; and Li, J. 2025.
\newblock GraphArena: Evaluating and Improving Large Language Models on Graph Computation.
\newblock In \emph{International Conference on Learning Representations}.

\bibitem[{Vogiatzis and Pardalos(2013)}]{COlogistics}
Vogiatzis, C.; and Pardalos, P. 2013.
\newblock \emph{Combinatorial optimization in transportation and logistics networks}, volume 2-5, 673--722.
\newblock Germany: Springer.
\newblock ISBN 9781441979964.
\newblock Publisher Copyright: {\textcopyright} Springer Science+Business Media New York 2013. All rights are reserved.

\bibitem[{xAI(2025)}]{grok3}
xAI. 2025.
\newblock Grok-3 and the Next Phase of xAI.
\newblock \url{https://x.ai/news/grok-3}.
\newblock Accessed: 2025-03-24.

\bibitem[{Xiao et~al.(2024{\natexlab{a}})Xiao, Zhang, Wu, Xu, Wang, Han, Fu, Zhong, Zeng, Song, and Chen}]{Xiao2024ChainofExpertsWL}
Xiao, Z.; Zhang, D.; Wu, Y.; Xu, L.; Wang, Y.~J.; Han, X.; Fu, X.; Zhong, T.; Zeng, J.; Song, M.; and Chen, G. 2024{\natexlab{a}}.
\newblock Chain-of-Experts: When LLMs Meet Complex Operations Research Problems.
\newblock In \emph{International Conference on Learning Representations}.

\bibitem[{Xiao et~al.(2024{\natexlab{b}})Xiao, Zhang, Wu, Xu, Wang, Han, Fu, Zhong, Zeng, Song, and Chen}]{xiao2024chainofexperts}
Xiao, Z.; Zhang, D.; Wu, Y.; Xu, L.; Wang, Y.~J.; Han, X.; Fu, X.; Zhong, T.; Zeng, J.; Song, M.; and Chen, G. 2024{\natexlab{b}}.
\newblock Chain-of-Experts: When {LLM}s Meet Complex Operations Research Problems.
\newblock In \emph{The Twelfth International Conference on Learning Representations}.

\bibitem[{Yang et~al.(2025{\natexlab{a}})Yang, Wang, Huang, Guo, Shi, Han, Feng, Song, Liang, and Tang}]{YANG2024OptiBenchMR}
Yang, Z.; Wang, Y.; Huang, Y.; Guo, Z.; Shi, W.; Han, X.; Feng, L.; Song, L.; Liang, X.; and Tang, J. 2025{\natexlab{a}}.
\newblock OptiBench Meets ReSocratic: Measure and Improve {LLM}s for Optimization Modeling.
\newblock In \emph{The Thirteenth International Conference on Learning Representations}.

\bibitem[{Yang et~al.(2025{\natexlab{b}})Yang, Wang, Huang, Guo, Shi, Han, Feng, Song, Liang, and Tang}]{yang2025optibench}
Yang, Z.; Wang, Y.; Huang, Y.; Guo, Z.; Shi, W.; Han, X.; Feng, L.; Song, L.; Liang, X.; and Tang, J. 2025{\natexlab{b}}.
\newblock OptiBench Meets ReSocratic: Measure and Improve {LLM}s for Optimization Modeling.
\newblock In \emph{The Thirteenth International Conference on Learning Representations}.

\bibitem[{Yao et~al.(2022)Yao, Zhao, Yu, Du, Shafran, Narasimhan, and Cao}]{Yao2022ReActSR}
Yao, S.; Zhao, J.; Yu, D.; Du, N.; Shafran, I.; Narasimhan, K.; and Cao, Y. 2022.
\newblock ReAct: Synergizing Reasoning and Acting in Language Models.
\newblock \emph{ArXiv}, abs/2210.03629.

\bibitem[{Ye et~al.(2024)Ye, Wang, Cao, Berto, Hua, Kim, Park, and Song}]{ye2024reevo}
Ye, H.; Wang, J.; Cao, Z.; Berto, F.; Hua, C.; Kim, H.; Park, J.; and Song, G. 2024.
\newblock ReEvo: Large Language Models as Hyper-Heuristics with Reflective Evolution.
\newblock In \emph{The Thirty-eighth Annual Conference on Neural Information Processing Systems}.

\bibitem[{Zheng et~al.(2025)Zheng, Xie, Wang, and Hooi}]{Zheng2025MonteCT}
Zheng, Z.; Xie, Z.; Wang, Z.; and Hooi, B. 2025.
\newblock Monte Carlo Tree Search for Comprehensive Exploration in LLM-Based Automatic Heuristic Design.
\newblock \emph{ArXiv}, abs/2501.08603.

\end{thebibliography}

\clearpage

\appendix

\section{Problem Description and Scores}

\subsection{Aircraft landing}
The problem is to schedule landing times for a set of planes across one or more runways such that each landing occurs within its prescribed time window and all pairwise separation requirements are satisfied; specifically, if plane i lands at or before plane j on the same runway, then the gap between their landing times must be at least the specified separation time provided in the input. In a multiple-runway setting, each plane must also be assigned to one runway, and if planes land on different runways, the separation requirement (which may differ) is applied accordingly. Each plane has an earliest, target, and latest landing time, with penalties incurred proportionally for landing before (earliness) or after (lateness) its target time. The objective is to minimize the total penalty cost while ensuring that no constraints are violated—if any constraint is breached, the solution receives no score.
\begin{table}[h]
\begin{tabular}{l cccc}
\toprule
Method & Score \\
\midrule

Classical Solver  &  0.5985295365478638 \\
BestOfN  &  0.8057479826999232 \\
Refine  &  0.7503157815146175 \\
FunSearch  &  0.9688863336568327 \\
AIDE  &  0.800637046201484 \\
ReEvo  &  0.9134454710810906 \\
MCTS  &  0.801655240273729 \\
EoH  &  0.8019818529389835 \\
\bottomrule
\end{tabular}
\caption{Aircraft landing}
\end{table}

\subsection{Assignment problem}
The Assignment Problem involves optimally assigning  n  items to  n  agents based on a provided  n 	imes n  cost matrix, where each entry $ext{cost\_matrix}[i][j]$ denotes the cost of assigning item  i+1  to agent  j+1 . The goal is to identify a permutation—each item assigned exactly one agent—that minimizes the total assignment cost. Formally, this is an optimization problem to find a permutation $\pi$ of agents such that the total cost $\sum{i=1}^{n} 	ext{cost\_matrix}[i-1][\pi(i)-1]$ is minimized. The solution returned includes both the minimal total cost and the corresponding optimal assignments.

\begin{table}[h]
\begin{tabular}{l cccc}
\toprule
Method & Score \\
\midrule

Classical Solver  &  1 \\
BestOfN  &  1 \\
Refine  &  1 \\
FunSearch  &  1 \\
AIDE  &  1 \\
ReEvo  &  1 \\
MCTS  &  1 \\
EoH  &  1 \\
\bottomrule
\end{tabular}
\caption{Assignment problem}
\end{table}

\subsection{Assortment problem}
This optimization problem involves arranging a set of rectangular pieces within available stock rectangles to minimize the overall waste area percentage. Each stock rectangle has a defined area, and each piece—which may be rotated by 90°—must be fully contained within a stock without overlapping with other pieces. Additionally, each piece type has specific total minimum and maximum placement limits. You have access to an unlimited number of stocks for each type, but you may use at most two stock types. The objective is to achieve the lowest possible waste area percentage, defined as the ratio of unused area to the total stock area. Solutions must ensure efficient resource utilization while satisfying all geometric and quantity constraints. Any violation of these constraints results in no score.
\begin{table}[h]
\begin{tabular}{l cccc}
\toprule
Method & Score \\
\midrule

Classical Solver  &  0.3222852468406736 \\
BestOfN  &  0.36161788534475603 \\
Refine  &  0.10475936163370339 \\
FunSearch  &  0.3622886282031154 \\
AIDE  &  0.1698107561339298 \\
ReEvo  &  0.24290833308629933 \\
MCTS  &  0.1757439194813797 \\
EoH  &  0.2519474328966603 \\
\bottomrule
\end{tabular}
\caption{Assortment problem}
\end{table}

\subsection{Bin packing - one-dimensional}
The **one-dimensional bin packing problem** seeks to minimize the number of bins required to pack a given set of items while ensuring that the sum of item sizes within each bin does not exceed the specified bin capacity. Given a test case with an identifier (`id`), a fixed `bin\_capacity`, and a list of `num\_items` with their respective sizes (`items`), the objective is to find a packing arrangement that uses the least number of bins. The solution is evaluated based on the total `num\_bins` used, with invalid solutions (e.g., missing or duplicated items, or bins exceeding capacity) incurring a inf heavy penalty. The output must include the number of bins used and a valid assignment of item indices to bins.
\begin{table}[h]
\begin{tabular}{l cccc}
\toprule
Method & Score \\
\midrule

Classical Solver  &  0.9628049317089281 \\
BestOfN  &  0.8933315064694979 \\
Refine  &  0.9870315022407082 \\
FunSearch  &  0.9557154223933677 \\
AIDE  &  0.8366913237780297 \\
ReEvo  &  0.9492158360156572 \\
MCTS  &  0.9396436307329097 \\
EoH  &  0.9693475618912389 \\
\bottomrule
\end{tabular}
\caption{Bin packing - one-dimensional}
\end{table}

\subsection{Capacitated warehouse location}
The Capacitated Warehouse Location Problem with Splittable Demand aims to determine which warehouses to open and how to allocate portions of customer demands among these warehouses in order to minimize total costs. Given a set of potential warehouse locations, each with a fixed opening cost and capacity limit, and a set of customers with individual demands and associated per-unit assignment costs to each warehouse, the objective is to decide which warehouses to open and how to distribute each customer's demand among these open warehouses. The allocation must satisfy the constraint that the sum of portions assigned to each customer equals their total demand, and that the total demand allocated to any warehouse does not exceed its capacity. The optimization seeks to minimize the sum of fixed warehouse opening costs and the total per-unit assignment costs. However, if any solution violates these constraints (i.e., a customer’s demand is not fully satisfied or a warehouse’s capacity is exceeded), then no score is provided.
\begin{table}[h]
\begin{tabular}{l cccc}
\toprule
Method & Score \\
\midrule

Classical Solver  &  0.6976400141361688 \\
BestOfN  &  0.0 \\
Refine  &  0.7518838886310322 \\
FunSearch  &  0.7196713948459038 \\
AIDE  &  0.6647355906610447 \\
ReEvo  &  0.6715266955394039 \\
MCTS  &  0.6891495773105485 \\
EoH  &  0.7502493181324346 \\
\bottomrule
\end{tabular}
\caption{Capacitated warehouse location}
\end{table}

\subsection{Common due date scheduling}
Given \(floor \), where \( h \) is a predefined fraction (defaulting to 0.6). The goal is to determine an optimal job sequence that minimizes the penalty, calculated as follows: for each job, if its completion time \( C \) is earlier than \( d \), an earliness penalty of \( a 	imes (d - C) \) is incurred; if \( C \) exceeds \( d \), a tardiness penalty of \( b 	imes (C - d) \) is applied; otherwise, no penalty is incurred. The problem requires finding a permutation of job indices (1-based) that minimizes the total penalty. The evaluation metric sums these penalties for a given schedule.

\begin{table}[h]
\begin{tabular}{l cccc}
\toprule
Method & Score \\
\midrule

Classical Solver  &  0.9187662046144239 \\
BestOfN  &  0.97731110557282 \\
Refine  &  0.9776844987221935 \\
FunSearch  &  0.976604327923604 \\
AIDE  &  0.6291657473867996 \\
ReEvo  &  0.9743199070415761 \\
MCTS  &  0.8838457578182489 \\
EoH  &  0.9773286503168127 \\
\bottomrule
\end{tabular}
\caption{Common due date scheduling}
\end{table}

\subsection{Constrained guillotine cutting}
The problem involves optimizing the guillotine feasibl placement of a set of rectangular pieces on a given stock sheet to maximize total value. Each piece type is characterized by its length, width, an upper bound on the number of times it may appear in the final cutting pattern, and an assigned value. Orientation of the pieces is fixed (the edges of the pieces are parallel to the edges of the sheet). The task is to select and place pieces such that each lies completely within the boundaries of the stock sheet, no two pieces overlap, and the number of pieces of any type does not exceed its specified maximum. A set of placements is considered guillotine feasible if there exists at least one straight cut (vertical or horizontal) that does not slice through any rectangle, and the property holds recursively on the resulting subregions. Empty regions or regions exactly matching a placed piece are considered valid.The objective is to maximize the sum of the values of the placed pieces; however, if any spatial or count constraint is violated, the solution is deemed invalid. The output is defined as a dictionary reporting the total value and a list of placements, with each placement specified by the piece type index, x and y coordinates, placed dimensions, and orientation flag.
\begin{table}[h]
\begin{tabular}{l cccc}
\toprule
Method & Score \\
\midrule

Classical Solver  &  0.7844900098230463 \\
BestOfN  &  0.0 \\
Refine  &  0.981513704843915 \\
FunSearch  &  0.956424099109148 \\
AIDE  &  0.9102922923098641 \\
ReEvo  &  0.0 \\
MCTS  &  0.0 \\
EoH  &  0.0 \\
\bottomrule
\end{tabular}
\caption{Constrained guillotine cutting}
\end{table}

\subsection{Constrained non-guillotine cutting}
The constrained non-guillotine cutting problem involves optimally arranging rectangular pieces onto a single rectangular stock with fixed dimensions (stock\_length and stock\_width). Each piece type has defined length, width, value, and minimum and maximum usage constraints. The optimization goal is to maximize the total value of all placed pieces, subject to constraints that each piece is entirely within stock boundaries, pieces do not overlap, each piece type’s usage falls within its specified [min, max] range, and pieces may optionally be rotated by 90°. The solution returns a set of placements indicating piece type, bottom-left coordinates (x, y), and rotation status.  If any constraint is violated, the solution receives no score.
\begin{table}[h]
\begin{tabular}{l cccc}
\toprule
Method & Score \\
\midrule

Classical Solver  &  0.5585076432266227 \\
BestOfN  &  0.8760613343780126 \\
Refine  &  0.99138085452391 \\
FunSearch  &  0.9623447685846964 \\
AIDE  &  0.8555320134962818 \\
ReEvo  &  0.9264764236682984 \\
MCTS  &  0.7944732650186651 \\
EoH  &  0.9106930512513293 \\
\bottomrule
\end{tabular}
\caption{Constrained non-guillotine cutting}
\end{table}

\subsection{Container loading}
Solves a container loading problem: Given a 3D container of specified dimensions and multiple box types—each defined by dimensions, orientation constraints, and available quantity—the goal is to optimally place these boxes within the container to maximize the volume utilization ratio. Each box placement must respect orientation constraints (vertical alignment flags), fit entirely within container boundaries, and avoid overlaps. The solution returns precise coordinates and orientations for each box placement, quantified by a volume utilization score calculated as the total volume of placed boxes divided by the container volume. Invalid placements result in a score of 0.0.
\begin{table}[h]
\begin{tabular}{l cccc}
\toprule
Method & Score \\
\midrule

Classical Solver  &  0.09700224776623062 \\
BestOfN  &  0.8163545342051534 \\
Refine  &  0.18895711345505883 \\
FunSearch  &  0.23070987019597894 \\
AIDE  &  0.7592850816892841 \\
ReEvo  &  0.716081346719743 \\
MCTS  &  0.5451472798828618 \\
EoH  &  0.7795824394970114 \\
\bottomrule
\end{tabular}
\caption{Container loading}
\end{table}

\subsection{Container loading with weight restrictions}
The Container Loading with Weight Restrictions problem aims to maximize the utilization of a container’s volume by selecting and strategically placing boxes inside it. Given a container with specified dimensions (length, width, height) and multiple types of boxes, each characterized by their dimensions, quantities, weights, and load-bearing constraints, the optimization goal is to determine the placement and orientation of these boxes (with each box allowed three possible orientations) that maximizes the ratio of total occupied box volume to container volume. The solution must strictly adhere to spatial constraints (boxes must fit entirely within the container without overlapping), load-bearing constraints (boxes must support the weight of boxes stacked above them according to given limits), and orientation restrictions. The optimization quality is evaluated by the achieved utilization metric, defined as the total volume of successfully placed boxes divided by the container volume; if any constraint is violated, the utilization score is zero.
\begin{table}[h]
\begin{tabular}{l cccc}
\toprule
Method & Score \\
\midrule

Classical Solver  &  0.009225308452359507 \\
BestOfN  &  0.13669723873453465 \\
Refine  &  0.07941319051933145 \\
FunSearch  &  0.2919729304847129 \\
AIDE  &  0.12860429344072807 \\
ReEvo  &  0.1420943670465572 \\
MCTS  &  0.04806324649022297 \\
EoH  &  0.051972410039456414 \\
\bottomrule
\end{tabular}
\caption{Container loading with weight restrictions}
\end{table}

\subsection{Corporate structuring}
Given N countries, each defined by:
  • a tax code (1: Exemption, 2: Deduction, 3: Source-by-source Pooling, 4: World-wide Pooling),
  • a foreign income tax rate,
  • a domestic income tax rate, and
  • a profit,
and a withholding tax matrix W (where W[i][j] is the rate on dividends from country i to j), construct a valid tree‐structured corporate hierarchy (directed, acyclic, connected) rooted at a designated target (whose parent is 0) such that every country with profit > 0 appears exactly once.

For each country i, define S as the set of nodes in its subtree (note the subtree includes itself) with a positive profit. Also consider the set of child nodes C\_i. 
If i is not a root country but in the tree, it will send all its income (after tax) to its parent j. Denote this amount as F[i][j]. Assume the total income after domestic tax and withholding tax for country i is: $domestic_income_i*(1-domestic_rate_i) + (\sum_{k\in C_i} F[k][i]*(1-W[k][i]))$
The extra foreign tax under different tax code is defined as follows:
    1. No extra tax.
    2. Foreign income tax from the child nodes: $foreign_income_rate_i*(\sum_{k\in C_i} F[k][i]*(1-W[k][i]))$
    3. Foreign income tax computed from the source nodes in each child's subtree: $\sum_{k\in C_i} max(0, F[k][i]*(1-W[k][i]) - (1-foreign_income_rate_i)*(\sum_{s\in S_k} domestic_income_s))$
    4. Foreign income tax from all source nodes in the subtree, excluding itself: max(0, $\sum_{k\in C_i}  F[k][i]*(1-W[k][i]) - (1-foreign_income_rate_i)*((\sum_{s\in S_i} domestic_income_s)-domestic_income_i))$
\begin{table}[h]
\begin{tabular}{l cccc}
\toprule
Method & Score \\
\midrule

Classical Solver  &  0.9450572839481785 \\
BestOfN  &  0.9450572839481785 \\
Refine  &  0.9726337326585759 \\
FunSearch  &  0.777775452943618 \\
AIDE  &  0.9450572839481785 \\
ReEvo  &  0.5014939649568603 \\
MCTS  &  0.9844897288603699 \\
EoH  &  0.9431107030735252 \\
\bottomrule
\end{tabular}
\caption{Corporate structuring}
\end{table}

\subsection{Crew scheduling}
The Crew Scheduling Problem involves assigning each task—with defined start and finish times—to exactly one crew, aiming to minimize the total transition costs between consecutive tasks. Each crew’s schedule must satisfy three constraints: tasks within a crew must not overlap; valid transitions (with associated costs) must exist between every consecutive pair of tasks; and the crew’s total duty time (from the start of the first task to the finish of the last) cannot exceed a specified time limit. Additionally, no more than K crews can be used to cover all tasks. Solutions violating any of these constraints are considered infeasible and receive no score. The optimization objective is therefore to determine assignments of tasks to no more than K crews that minimize the sum of transition costs while strictly adhering to all constraints, yielding a feasible and cost-effective scheduling solution.
\begin{table}[h]
\begin{tabular}{l cccc}
\toprule
Method & Score \\
\midrule

Classical Solver  &  0.45498811952880935 \\
BestOfN  &  0.4483461488661745 \\
Refine  &  0.6690343590115082 \\
FunSearch  &  0.5536756258756895 \\
AIDE  &  0.44095505708697136 \\
ReEvo  &  0.45225267224663634 \\
MCTS  &  0.4446817469828879 \\
EoH  &  0.5864457661923881 \\
\bottomrule
\end{tabular}
\caption{Crew scheduling}
\end{table}

\subsection{Equitable partitioning problem}
The task is to partition a set of individuals—each characterized by multiple binary attributes—into exactly 8 groups such that the distribution of attribute values is as balanced as possible across these groups. For each attribute, count the number of individuals with a ‘1’ in each group. The optimization objective is to minimize the total imbalance, which is defined as follows: for each attribute, calculate the absolute differences between the count in each group and the mean count across all groups, take the average of these differences, and then sum these averages over all attributes. The goal is to determine a group assignment for each individual that achieves the lowest possible total imbalance score.
\begin{table}[h]
\begin{tabular}{l cccc}
\toprule
Method & Score \\
\midrule

Classical Solver  &  1.0 \\
BestOfN  &  1.0 \\
Refine  &  1.0 \\
FunSearch  &  1.0 \\
AIDE  &  0.7777777777777778 \\
ReEvo  &  0.7777777777777778 \\
MCTS  &  1.0 \\
EoH  &  1.0 \\
\bottomrule
\end{tabular}
\caption{Equitable partitioning problem}
\end{table}

\subsection{Euclidean Steiner problem}
Given a set of 2D points (terminals), the goal of the Euclidean Steiner Problem is to compute a tree connecting all terminals with minimum total length. The total length is measured as the sum of Euclidean distances (where the Euclidean distance between two points $(x1,y1)$ and $(x2,y2)$ is $sqrt((x1-x2)^2 + (y1-y2)^2))$. Unlike a Minimum Spanning Tree (MST) computed solely on the given terminals, a Steiner tree may introduce extra points, called Steiner points, to reduce the overall length. In this formulation, it is assumed that the final candidate tree’s total length is given by the MST computed on the union of the original terminals and the reported Steiner points. A lower ratio (candidate\_tree\_length/MST\_original\_length) indicates a better solution.
\begin{table}[h]
\begin{tabular}{l cccc}
\toprule
Method & Score \\
\midrule

Classical Solver  &  0.9779703480188361 \\
BestOfN  &  0.6291391910535526 \\
Refine  &  0.688025642110573 \\
FunSearch  &  0.6968176110449371 \\
AIDE  &  0.04483890014026932 \\
ReEvo  &  0.5469067768233761 \\
MCTS  &  0.43093954323065975 \\
EoH  &  0.5917817000598826 \\
\bottomrule
\end{tabular}
\caption{Euclidean Steiner problem}
\end{table}

\subsection{Flow shop scheduling}
Given  n  jobs and  m  machines, the goal of the flow shop scheduling problem is to determine the optimal job sequence that minimizes the makespan, i.e., the total time required to complete all jobs on all machines. Each job follows the same machine order, and the processing times are specified in an  n 	imes m  matrix. The output is a permutation of job indices representing the processing order. If the constraints are not satisfied (e.g., invalid job sequencing), the solution receives no score. The objective is to optimize the makespan using the classical flow shop recurrence.
\begin{table}[h]
\begin{tabular}{l cccc}
\toprule
Method & Score \\
\midrule

Classical Solver  &  0.9222700445897257 \\
BestOfN  &  0.874217493803887 \\
Refine  &  0.8463439348165006 \\
FunSearch  &  0.8537338049420798 \\
AIDE  &  0.9144895115672386 \\
ReEvo  &  0.8424667927400846 \\
MCTS  &  0.9242143967817102 \\
EoH  &  0.940154419652199 \\
\bottomrule
\end{tabular}
\caption{Flow shop scheduling}
\end{table}

\subsection{Generalised assignment problem}
Generalized Assignment Problem (GAP)

The Generalized Assignment Problem (GAP) involves assigning \( n \) jobs to \( m \) agents such that each job is assigned to exactly one agent, and the resource consumption for each agent does not exceed its capacity. The objective is to optimize the total cost based on the problem type. When formulated as a maximization problem, the goal is to maximize the total cost; when formulated as a minimization problem, the goal is to minimize the total cost. Given a cost matrix (representing the cost of assigning jobs to agents), a consumption matrix (indicating the resource usage per assignment), and capacities (the resource limits for each agent), the task is to find a valid assignment that meets the capacity constraints while optimizing the total cost as specified by the problem indicator.

\begin{table}[h]
\begin{tabular}{l cccc}
\toprule
Method & Score \\
\midrule

Classical Solver  &  1.000509368510793 \\
BestOfN  &  1.000152715871272 \\
Refine  &  0.9997973477884884 \\
FunSearch  &  0.9360910283983461 \\
AIDE  &  1.000152715871272 \\
ReEvo  &  1.0002083856508814 \\
MCTS  &  1.0001026538510593 \\
EoH  &  0.9793902133221158 \\
\bottomrule
\end{tabular}
\caption{Generalised assignment problem}
\end{table}

\subsection{Graph colouring}
Given a graph in DIMACS format with vertices, edges, and an adjacency list, the goal is to assign a positive integer color (1..n) to each vertex while ensuring that no two adjacent vertices share the same color. The objective is to minimize the number of distinct colors used. If any two adjacent vertices have the same color, the solution is invalid and receives no score. Otherwise, the score is equal to the number of distinct colors used, with a lower score being better.
\begin{table}[h]
\begin{tabular}{l cccc}
\toprule
Method & Score \\
\midrule

Classical Solver  &  0.8679121232535366 \\
BestOfN  &  0.7992347794550977 \\
Refine  &  0.9237393162393163 \\
FunSearch  &  0.8993461774953884 \\
AIDE  &  0.7992347794550977 \\
ReEvo  &  0.8119485901255648 \\
MCTS  &  0.8529682767415909 \\
EoH  &  0.804175457505431 \\
\bottomrule
\end{tabular}
\caption{Graph colouring}
\end{table}

\subsection{Hybrid Reentrant Shop Scheduling}
The problem is a Hybrid Reentrant Shop Scheduling problem where each of n jobs must sequentially undergo three operations: an initialization phase on one of m identical primary machines, a setup phase on a single remote server, and a final main processing phase on the same primary machine used for initialization. Jobs are initialized in a fixed natural order using list scheduling, while the setup phase is processed on the remote server in an order specified by a permutation decision variable. Additionally, each job is assigned to a primary machine for main processing via a batch\_assignment, and on each machine, jobs are processed in natural (initialization) order. The objective is to minimize the makespan, defined as the time when the last job completes its main processing, while ensuring that no machine (primary or server) processes more than one job simultaneously and that all operational precedence constraints are satisfied.
\begin{table}[h]
\begin{tabular}{l cccc}
\toprule
Method & Score \\
\midrule

Classical Solver  &  0.9057971372430776 \\
BestOfN  &  0.9872450518587456 \\
Refine  &  0.9966666343001128 \\
FunSearch  &  1.0001780484032463 \\
AIDE  &  0.7457203947696327 \\
ReEvo  &  0.9820554515396009 \\
MCTS  &  0.9961239866411462 \\
EoH  &  0.9841146688046011 \\
\bottomrule
\end{tabular}
\caption{Hybrid Reentrant Shop Scheduling}
\end{table}

\subsection{Job shop scheduling}
The job shop scheduling problem requires assigning non-negative integer start times to a set of operations, structured into multiple jobs, each composed of sequential operations. Each operation is processed on a specific machine for a given processing time. The optimization goal is to minimize the makespan, defined as the maximum completion time across all jobs. Constraints include (i) sequential processing of operations within each job, meaning each operation cannot start before its preceding operation finishes, and (ii) non-overlapping scheduling of operations on the same machine. If these constraints are violated, the solution receives no score.
\begin{table}[h]
\begin{tabular}{l cccc}
\toprule
Method & Score \\
\midrule

Classical Solver  &  0.8202016779421567 \\
BestOfN  &  0.7060712883377539 \\
Refine  &  0.7696287350855926 \\
FunSearch  &  0.8192815531664928 \\
AIDE  &  0.6498336005961379 \\
ReEvo  &  0.7982807066317813 \\
MCTS  &  0.7293663754433233 \\
EoH  &  0.7770594374788831 \\
\bottomrule
\end{tabular}
\caption{Job shop scheduling}
\end{table}

\subsection{MIS}
The Maximum Independent Set (MIS) problem is a fundamental NP-hard optimization problem in graph theory. Given an undirected graph G = (V, E), where V is a set of vertices and E is a set of edges, the goal is to find the largest subset S in V such that no two vertices in S are adjacent (i.e., connected by an edge).
\begin{table}[h]
\begin{tabular}{l cccc}
\toprule
Method & Score \\
\midrule

Classical Solver  &  0.986 \\
BestOfN  &  0.8461150261004076 \\
Refine  &  0.9078324503859446 \\
FunSearch  &  0.9002038932676987 \\
AIDE  &  0.8425484500134511 \\
ReEvo  &  0.8342509729450779 \\
MCTS  &  0.8433127163177989 \\
EoH  &  0.8763795109859694 \\
\bottomrule
\end{tabular}
\caption{MIS}
\end{table}

\subsection{Multi-Demand Multidimensional Knapsack problem}
The Multi-Demand Multidimensional Knapsack Problem (MDMKP) is a binary optimization problem that extends the classical MKP by incorporating both upper-bound (<=) and lower-bound (>=) constraints. Formally, given n decision variables $x_j \in \{0,1\}$, the goal is to maximize $\sum_{j=1}^n c_j x_j$ subject to$ \sum_{j=1}^n a_{ij} x_j \le b_i for i=1,\dots,m$ and $\sum_{j=1}^n a_{ij} x_j \ge b_i for i=m+1,\dots,m+q$. Instances are generated from standard MKP problems by varying the number of >= constraints (with q taking values 1, m/2, or m) and by using two types of cost coefficients (positive and mixed), thereby producing six distinct variants per base instance. This formulation enables rigorous evaluation of algorithms in contexts where both resource limits and demand fulfillment must be simultaneously addressed.
\begin{table}[h]
\begin{tabular}{l cccc}
\toprule
Method & Score \\
\midrule

Classical Solver  &  0.8957822313136857 \\
BestOfN  &  0.7144432351611377 \\
Refine  &  0.8913402342031996 \\
FunSearch  &  0.8354799525874899 \\
AIDE  &  0.8805432369541204 \\
ReEvo  &  0.8920786376031828 \\
MCTS  &  0.8994648109682947 \\
EoH  &  0.9082814870567889 \\
\bottomrule
\end{tabular}
\caption{Multi-Demand Multidimensional Knapsack problem}
\end{table}

\subsection{Multidimensional knapsack problem}
This problem is a multidimensional knapsack optimization where the objective is to maximize the total profit by selecting decision variables, each associated with a profit and resource consumption across multiple constraints. The decision variables must be chosen such that the sum of resource usage for each constraint does not exceed its corresponding capacity. Importantly, if any constraint is violated—that is, if the resource consumption for any constraint exceeds its allowed capacity—the solution is deemed infeasible and earns no score. The challenge lies in identifying the optimal combination of items that yields the highest total profit while strictly satisfying all resource constraints.
\begin{table}[h]
\begin{tabular}{l cccc}
\toprule
Method & Score \\
\midrule

Classical Solver  &  0.9903523477639424 \\
BestOfN  &  0.9401685100749627 \\
Refine  &  0.9947726903727786 \\
FunSearch  &  0.9773347714972982 \\
AIDE  &  0.925117898068383 \\
ReEvo  &  1.0018885951740353 \\
MCTS  &  1.0057751617808324 \\
EoH  &  1.0010112897238341 \\
\bottomrule
\end{tabular}
\caption{Multidimensional knapsack problem}
\end{table}

\subsection{Open shop scheduling}
The Open Shop Scheduling Problem involves scheduling a set of jobs across a set of machines with the goal of minimizing the total completion time (makespan). Each job consists of several operations, where each operation must be processed on a specific machine for a given duration. Unlike other scheduling problems, the Open Shop variant has no predetermined order for processing the operations of a job—operations can be scheduled in any order, but a job can only be processed on one machine at a time, and a machine can only process one job at a time. This creates a complex combinatorial optimization challenge where the scheduler must determine both the sequence of operations for each job and the timing of each operation to minimize the overall completion time while ensuring no resource conflicts.
\begin{table}[h]
\begin{tabular}{l cccc}
\toprule
Method & Score \\
\midrule

Classical Solver  &  0.7851209868863173 \\
BestOfN  &  0.9017764948703829 \\
Refine  &  0.9930284498507208 \\
FunSearch  &  0.9930284498507208 \\
AIDE  &  0.9156437907474381 \\
ReEvo  &  0.9825099803205837 \\
MCTS  &  0.8960699709846601 \\
EoH  &  0.9930284498507208 \\
\bottomrule
\end{tabular}
\caption{Open shop scheduling}
\end{table}

\subsection{Packing unequal circles}
The problem involves packing a subset of unequal circles into a fixed circular container with radius R\_0 and center at the origin, where each circle i has a given radius R\_i (sorted in non-decreasing order) and is associated with a binary decision variable alpha\_i indicating whether it is packed. The goal is to maximize the number of circles packed—that is, maximize $\sum_{i=1}^{n}\alpha_i$—subject to two sets of nonlinear constraints: (1) each packed circle must lie entirely within the container, which is enforced by ensuring that the distance from its center to the container’s center plus its radius does not exceed R\_0; and (2) any two packed circles must not overlap, meaning the distance between their centers must be at least the sum of their radii.
\begin{table}[h]
\begin{tabular}{l cccc}
\toprule
Method & Score \\
\midrule

Classical Solver  &  0.9075757575757577 \\
BestOfN  &  0.8939393939393939 \\
Refine  &  0.9803030303030303 \\
FunSearch  &  0.9719696969696969 \\
AIDE  &  0.8825757575757576 \\
ReEvo  &  0.8825757575757576 \\
MCTS  &  0.9522727272727273 \\
EoH  &  0.8825757575757576 \\
\bottomrule
\end{tabular}
\caption{Packing unequal circles}
\end{table}

\subsection{Packing unequal circles area}
The problem involves packing a subset of unequal circles into a fixed circular container with radius R\_0 and center at the origin, where each circle i has a given radius R\_i (sorted in non-decreasing order) and is associated with a binary decision variable alpha\_i indicating whether it is packed. The goal is to maximize the total area of all circles packed—that is, maximize $\sum_{i=1}^{n}\alpha_i*pi*R_i^2$—subject to two sets of nonlinear constraints: (1) each packed circle must lie entirely within the container, which is enforced by ensuring that the distance from its center to the container’s center plus its radius does not exceed R\_0; and (2) any two packed circles must not overlap, meaning the distance between their centers must be at least the sum of their radii.
\begin{table}[h]
\begin{tabular}{l cccc}
\toprule
Method & Score \\
\midrule

Classical Solver  &  0.8767896840297265 \\
BestOfN  &  0.9923476599194556 \\
Refine  &  1.0226692239919217 \\
FunSearch  &  1.0404725950195108 \\
AIDE  &  0.5972138868724692 \\
ReEvo  &  0.9101821460280035 \\
MCTS  &  0.9617483396206504 \\
EoH  &  1.0056059827170811 \\
\bottomrule
\end{tabular}
\caption{Packing unequal circles area}
\end{table}

\subsection{Packing unequal rectangles and squares}
We are given a set of n unequal rectangles (or squares), each with specified dimensions, and a fixed circular container of radius R centered at the origin. The problem is to decide which rectangles to pack and where to position them—by choosing binary selection variables and continuous center coordinates—so that every packed rectangle is entirely contained within the circle and no two packed rectangles overlap. For each rectangle, the four corners must lie inside the circle, and if an item is not packed it is forced to a dummy position. The objective is to maximize the number of packed items, i.e., maximize $\sum_{i=1}^{n} alpha_i$ (or a related sum when each alpha\_i is binary). Note that the rotation of the rectagular (by 90 degrees) is sometimes allowed and your algorithm should take that into account.
\begin{table}[h]
\begin{tabular}{l cccc}
\toprule
Method & Score \\
\midrule

Classical Solver  &  0.9134625513058007 \\
BestOfN  &  0.8337025039542202 \\
Refine  &  0.932172162950195 \\
FunSearch  &  0.9228828411608733 \\
AIDE  &  0.7950708457573447 \\
ReEvo  &  0.77954425754769 \\
MCTS  &  0.8028450160315149 \\
EoH  &  0.9228828411608733 \\
\bottomrule
\end{tabular}
\caption{Packing unequal rectangles and squares}
\end{table}

\subsection{Packing unequal rectangles and squares area}
We consider the problem of selecting and placing a subset of  n  unequal rectangles (or squares) into a fixed‐size circular container of radius  R  so as to maximize the total area of the packed items. Each item  i  has given dimensions  $L_i$  and  $W_i$  (with  $L_i = W_i $ for squares) and an associated area  $L_iW_i$ . The decision variables include a binary indicator $\alpha_i$ for whether item  i  is packed and continuous variables $(x_i, y_i)$ for the placement of its center, along with a rotation angle  	$heta_i$  when $90^\circ$ rotations are allowed. The formulation enforces that for every packed item, all four of its rotated corners must lie within the circle, and that no two packed items overlap; if an item is not packed, it is fixed at a dummy position.
\begin{table}[h]
\begin{tabular}{l cccc}
\toprule
Method & Score \\
\midrule

Classical Solver  &  0.8893527400499813 \\
BestOfN  &  0.9536806816195774 \\
Refine  &  1.0513451711752306 \\
FunSearch  &  1.0839011538182066 \\
AIDE  &  0.8100272732450019 \\
ReEvo  &  0.9435059488868657 \\
MCTS  &  0.995946490673633 \\
EoH  &  0.9566331174271511 \\
\bottomrule
\end{tabular}
\caption{Packing unequal rectangles and squares area}
\end{table}

\subsection{Resource constrained shortest path}
This problem involves finding the shortest path from vertex 1 to vertex n in a directed graph while satisfying resource constraints. Specifically, each vertex and arc has associated resource consumptions, and the cumulative consumption for each resource must fall within the provided lower\_bounds and upper\_bounds. The input includes the number of vertices (n), arcs (m), resource types (K), resource consumption at each vertex, and a graph represented as a mapping from vertices to lists of arcs (each arc being a tuple of end vertex, cost, and arc resource consumptions). The optimization objective is to minimize the total arc cost of the path, with the condition that the path is valid—meaning it starts at vertex 1, ends at vertex n, follows defined transitions in the graph, and respects all resource bounds; if any of these constraints are not met, the solution receives no score.
\begin{table}[h]
\begin{tabular}{l cccc}
\toprule
Method & Score \\
\midrule

Classical Solver  &  0.7508899529136809 \\
BestOfN  &  0.7508899529136808 \\
Refine  &  0.7284494767232047 \\
FunSearch  &  0.7508899529136808 \\
AIDE  &  0.7508899529136808 \\
ReEvo  &  0.7508899529136808 \\
MCTS  &  0.7284494767232047 \\
EoH  &  0.7508899529136808 \\
\bottomrule
\end{tabular}
\caption{Resource constrained shortest path}
\end{table}

\subsection{Set covering}
Set Covering Problem. The goal is to select a subset of columns, each with an associated cost, such that every row is covered by at least one chosen column. For each row, the available covering columns are provided (as 1-indexed numbers). The objective is to minimize the total cost of the selected columns, and if even one row is left uncovered, then no score is awarded.
\begin{table}[h]
\begin{tabular}{l cccc}
\toprule
Method & Score \\
\midrule

Classical Solver  &  0.8883906244045974 \\
BestOfN  &  0.8213286754887226 \\
Refine  &  0.9056204467263304 \\
FunSearch  &  0.8887733963981322 \\
AIDE  &  0.8639998129016312 \\
ReEvo  &  0.9360686599803572 \\
MCTS  &  0.8672991644233662 \\
EoH  &  0.8843920544743958 \\
\bottomrule
\end{tabular}
\caption{Set covering}
\end{table}

\subsection{Set partitioning}
This problem involves solving a set partitioning instance where the goal is to choose a subset of columns such that each row is covered exactly once while minimizing the total cost. Each column is associated with a cost and covers a specific set of rows. The optimization problem is defined by selecting columns from a given set so that every row is covered precisely once, and the sum of the selected columns’ costs is minimized. If the solution fails to cover every row exactly once, then no score is awarded.
\begin{table}[h]
\begin{tabular}{l cccc}
\toprule
Method & Score \\
\midrule

Classical Solver  &  0.9996401983661346 \\
BestOfN  &  0.8991338255841825 \\
Refine  &  0.7999991398515384 \\
FunSearch  &  0.8333333333333334 \\
AIDE  &  0.9 \\
ReEvo  &  0.8991338255841825 \\
MCTS  &  0.8647769492523454 \\
EoH  &  0.9324671589175159 \\
\bottomrule
\end{tabular}
\caption{Set partitioning}
\end{table}

\subsection{TSP}
The Traveling Salesman Problem (TSP) is a classic combinatorial optimization problem where, given a set of cities with known pairwise distances, the objective is to find the shortest possible tour that visits each city exactly once and returns to the starting city. More formally, given a complete graph G = (V, E) with vertices V representing cities and edges E with weights representing distances, we seek to find a Hamiltonian cycle (a closed path visiting each vertex exactly once) of minimum total weight.
\begin{table}[h]
\begin{tabular}{l cccc}
\toprule
Method & Score \\
\midrule

Classical Solver  &  0.986 \\
BestOfN  &  0.8590303340408165 \\
Refine  &  0.9399577646813952 \\
FunSearch  &  0.9016741050908584 \\
AIDE  &  0.7710495444635409 \\
ReEvo  &  0.8488918718349553 \\
MCTS  &  0.5961113158302597 \\
EoH  &  0.7935463156320405 \\
\bottomrule
\end{tabular}
\caption{TSP}
\end{table}

\subsection{Uncapacitated warehouse location}
The Uncapacitated Warehouse Location Problem aims to determine which warehouses to open and how to assign each customer entirely to an open warehouse in order to minimize the total cost. Given a set of potential warehouse locations, each with a fixed opening cost, and a set of customers, each with an associated assignment cost for being served by each warehouse, the objective is to select a subset of warehouses to open and assign every customer completely to one of these open warehouses. The optimization minimizes the sum of fixed warehouse opening costs and the customer assignment costs. Each customer must be assigned to exactly one warehouse; if any customer is left unassigned or assigned to more than one warehouse, the solution is considered infeasible.
\begin{table}[h]
\begin{tabular}{l cccc}
\toprule
Method & Score \\
\midrule

Classical Solver  &  0.9968157833494645 \\
BestOfN  &  0.98931916166557 \\
Refine  &  1.0000000000002045 \\
FunSearch  &  0.9978398298062331 \\
AIDE  &  0.9994999857664043 \\
ReEvo  &  0.998083746641369 \\
MCTS  &  0.9951604598088827 \\
EoH  &  0.8749999999978142 \\
\bottomrule
\end{tabular}
\caption{Uncapacitated warehouse location}
\end{table}

\subsection{Unconstrained guillotine cutting}
The unconstrained guillotine cutting problem involves selecting and placing a subset of available pieces within a fixed stock rectangle to maximize the total value of the placed pieces. Each piece, defined by its length, width, and value, may be optionally rotated 90° if allowed and used at most once. The challenge is to determine both the selection and the positioning of these pieces such that they do not overlap and lie entirely within the stock’s boundaries. This optimization problem formalizes the decision variables as the x and y coordinates for the bottom-left placement of each piece and, if rotation is allowed, a binary variable indicating its orientation, while the objective function is to maximize the sum of the values of the pieces successfully placed within the stock.
\begin{table}[h]
\begin{tabular}{l cccc}
\toprule
Method & Score \\
\midrule

Classical Solver  &  0.9725381370960237 \\
BestOfN  &  0.8701275303357732 \\
Refine  &  0.9618177725501762 \\
FunSearch  &  0.9646369625362231 \\
AIDE  &  0.8512970128354943 \\
ReEvo  &  0.9828452190272524 \\
MCTS  &  0.8628525304460628 \\
EoH  &  0.9649480933563296 \\
\bottomrule
\end{tabular}
\caption{Unconstrained guillotine cutting}
\end{table}

\subsection{Vehicle routing: period routing}
The Period Vehicle Routing Problem requires planning delivery routes over a multi‐day planning period.

Each customer (other than the depot, whose id is 0) is provided with a list of candidate service schedules. A schedule is represented by a binary vector of length equal to the period (e.g., [1, 0, 1] for a 3‐day period), where a 1 in a given position indicates that the customer must be visited on that day. The decision maker must select exactly one candidate schedule for each customer.

For every day in the planning period, if a customer’s chosen schedule indicates a delivery (i.e., a 1),
then exactly one vehicle must visit that customer on that day. Otherwise, the customer should not be visited. The decision maker must also design, for each day, the tours for the vehicles. Each tour is a continuous route that starts at the depot (id 0) and, after visiting a subset of customers, returns to the depot. Each vehicle is only allowed to visit the depot once per day—namely, as its starting and ending point—and it is not allowed to return to the depot in the middle of a tour.

Moreover, each vehicle route must obey a capacity constraint: the total demand of the customers visited on that tour must not exceed the vehicle capacity each day. Although multiple vehicles are available per day (as specified by the input), not all available vehicles have to be used, but the number of tours in a given day cannot exceed the provided number of vehicles. In addition, the tours on each day must cover exactly those customers who require service per the selected schedules, and no customer may be visited more than once in a given day.

The objective is to choose a schedule for every customer and plan the daily tours so as to minimize the overall distance traveled
by all vehicles during the entire planning period. Distances are measured using Euclidean distance.
\begin{table}[h]
\begin{tabular}{l cccc}
\toprule
Method & Score \\
\midrule

Classical Solver  &  0.12437943290991642 \\
BestOfN  &  0.42032326191804853 \\
Refine  &  0.48371172427664344 \\
FunSearch  &  0.32385035648314586 \\
AIDE  &  0.5362363612554435 \\
ReEvo  &  0.0 \\
MCTS  &  0.0 \\
EoH  &  0.0 \\
\bottomrule
\end{tabular}
\caption{Vehicle routing: period routing}
\end{table}

\subsection{p-median - capacitated}
The Capacitated P-Median Problem is a facility location optimization problem where the objective is to select exactly  p  customers as medians (facility locations) and assign each customer to one of these medians to minimize the total cost, defined as the sum of the Euclidean distances (rounded down to the nearest integer) between customers and their assigned medians. Each median has a capacity constraint  Q , meaning the total demand of the customers assigned to it cannot exceed  Q . A feasible solution must respect this capacity constraint for all medians; otherwise, it receives a score of zero. The solution is evaluated by the ratio  	ext{score} = rac{	ext{best\_known}}{	ext{computed\_total\_cost}} , where computed\_total\_cost is the total assignment cost if all constraints are satisfied; otherwise, the score is zero. The output consists of the total cost (if feasible), the selected medians, and the customer assignments.
\begin{table}[h]
\begin{tabular}{l cccc}
\toprule
Method & Score \\
\midrule

Classical Solver  &  0.8996179560649475 \\
BestOfN  &  0.9892886172082498 \\
Refine  &  0.9737771618997864 \\
FunSearch  &  0.9748437166838722 \\
AIDE  &  0.7442228395960961 \\
ReEvo  &  0.9786585768154689 \\
MCTS  &  0.9829650705934849 \\
EoH  &  0.9853458094532425 \\
\bottomrule
\end{tabular}
\caption{p-median - capacitated}
\end{table}

\subsection{p-median - uncapacitated}
The uncapacitated p-median problem is a combinatorial optimization problem defined on a given graph  G = (V, E)  with  n  vertices and  m  edges. The objective is to select  p  medians (facility locations) from the set of vertices such that the total assignment cost is minimized. The assignment cost is computed as the sum of the shortest distances from each vertex to its nearest selected median, where distances are given by a precomputed complete cost matrix (obtained via Floyd’s algorithm). Formally, given the cost matrix $ D \in \mathbb{R}^{n 	\times n}$ , the optimization problem seeks to find a subset  $S \subseteq V  with  |S| = p$  that minimizes the function:

$\sum_{v \in V} \min_{s \in S} D(v, s)$

where  D(v, s)  is the shortest-path distance between vertex  v  and median  s . The solution consists of a list of exactly  p  distinct vertices representing the chosen medians.
\begin{table}[h]
\begin{tabular}{l cccc}
\toprule
Method & Score \\
\midrule

Classical Solver  &  0.9952341868141825 \\
BestOfN  &  0.9453613019698086 \\
Refine  &  0.9982141349797949 \\
FunSearch  &  0.9996783954983718 \\
AIDE  &  0.9847816841274486 \\
ReEvo  &  0.9983315585722753 \\
MCTS  &  0.9605290267584901 \\
EoH  &  0.9921177098573016 \\
\bottomrule
\end{tabular}
\caption{p-median - uncapacitated}
\end{table}

% \section*{Broader Impact \& Ethics Statement}

% CO-Bench aims to advance the study of LLM-based agents in combinatorial optimization by providing a diverse and rigorous benchmark. By evaluating LLM agents on real-world CO problems, this work contributes to understanding their potential to automate and accelerate algorithm design in domains such as logistics, manufacturing, and scientific computing.

% We acknowledge that the use of LLM agents in decision-making systems—especially those deployed in high-stakes contexts—requires careful validation to ensure safety, fairness, and accountability. Our benchmark does not currently evaluate social or ethical impacts of discovered algorithms (e.g., fairness or resource allocation bias), which is an important area for future extension. Additionally, as LLMs may replicate biases from training data or generate unverified solutions, we caution against overreliance on these models without human oversight.
% By releasing CO-Bench and our findings, we aim to foster transparent, reproducible research and encourage the development of responsible, trustworthy LLM agents for optimization tasks.

% \input{ReproducibilityChecklist.tex}

\end{document}